\documentclass[sigconf, nonacm]{acmart}

\usepackage{adjustbox}
\usepackage{amsmath}
\usepackage{bbm}
\usepackage{xcolor}
\usepackage{url}
\usepackage{caption}
\usepackage[subrefformat=parens,labelformat=parens]{subcaption}
\usepackage[capitalise]{cleveref}

\usepackage{tikz}
\usetikzlibrary{positioning, arrows, arrows.meta, fit, calc}

\AtBeginDocument{%
  \providecommand\BibTeX{{%
    \normalfont B\kern-0.5em{\scshape i\kern-0.25em b}\kern-0.8em\TeX}}}

\setcopyright{acmcopyright}
\copyrightyear{2022}
\acmYear{2022}
\acmDOI{XXXXXXX.XXXXXXX}

\acmConference[SIGSPATIAL '22]{30th ACM SIGSPATIAL International Conference on Advances in Geographic Information Systems}{November 1--4, 2022}{Seattle, WA}
\acmPrice{15.00}
\acmISBN{978-1-4503-XXXX-X/18/06}

\begin{document}

\title{Learning Citywide Patterns of Life from Trajectory Monitoring}

\author{Mark Tenzer}
\email{mtenzer@novateur.ai}
\affiliation{%
  \institution{Novateur Research Solutions}
  \streetaddress{20110 Ashbrook Place, Suite 275}
  \city{Ashburn}
  \state{Virginia}
  \country{USA}
  \postcode{20147}
}

\author{Zeeshan Rasheed}
\email{zrasheed@novateur.ai}
\affiliation{%
  \institution{Novateur Research Solutions}
  \streetaddress{20110 Ashbrook Place, Suite 275}
  \city{Ashburn}
  \state{Virginia}
  \country{USA}
  \postcode{20147}
}

\author{Khurram Shafique}
\email{kshafique@novateur.ai}
\affiliation{%
  \institution{Novateur Research Solutions}
  \streetaddress{20110 Ashbrook Place, Suite 275}
  \city{Ashburn}
  \state{Virginia}
  \country{USA}
  \postcode{20147}
}

\renewcommand{\shortauthors}{Tenzer, Rasheed, and Shafique}

\begin{abstract}
The recent proliferation of real-world human mobility datasets has catalyzed geospatial and transportation research in trajectory prediction, demand forecasting, travel time estimation, and anomaly detection. However, these datasets also enable, more broadly, a descriptive analysis of intricate systems of human mobility. We formally define \emph{patterns of life analysis} as a natural, explainable extension of online unsupervised anomaly detection, where we not only monitor a data stream for anomalies but also explicitly extract normal patterns over time.  To learn patterns of life, we adapt Grow When Required (GWR) episodic memory from research in computational biology and neurorobotics to a new domain of geospatial analysis.  This biologically-inspired neural network, related to self-organizing maps (SOM), constructs a set of ``memories'' or prototype traffic patterns incrementally as it iterates over the GPS stream.  It then compares each new observation to its prior experiences, inducing an online, unsupervised clustering and anomaly detection on the data.  We mine patterns-of-interest from the Porto taxi dataset, including both major public holidays and newly-discovered transportation anomalies, such as festivals and concerts which, to our knowledge, have not been previously acknowledged or reported in prior work. We anticipate that the capability to incrementally learn normal and abnormal road transportation behavior will be useful in many domains, including smart cities, autonomous vehicles, and urban planning and management.
\end{abstract}



\begin{teaserfigure}
  \centering
  \adjustbox{width=\textwidth}{\begin{tikzpicture}

    \node[anchor=east, align=center, rectangle, draw=black, fill=white] (input) at (1.5,0) {GPS trajectory data \\ \includegraphics[width=5cm]{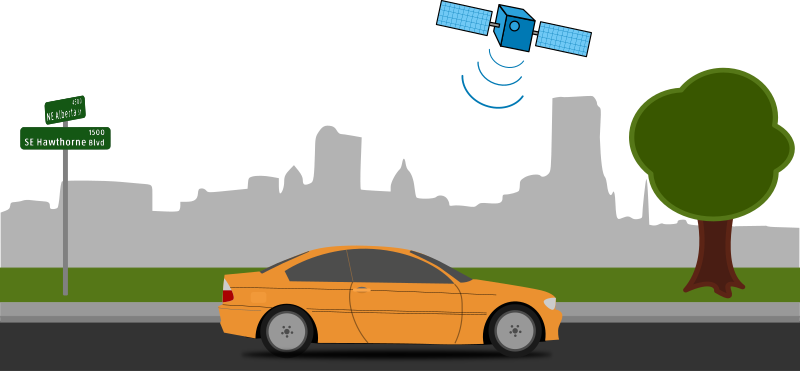}};
    
    \node[anchor=east, align=center, rectangle, draw=black] (region_gwr) at (-2,-3.5) {Region-level GWR \\ \includegraphics[height=2.5cm]{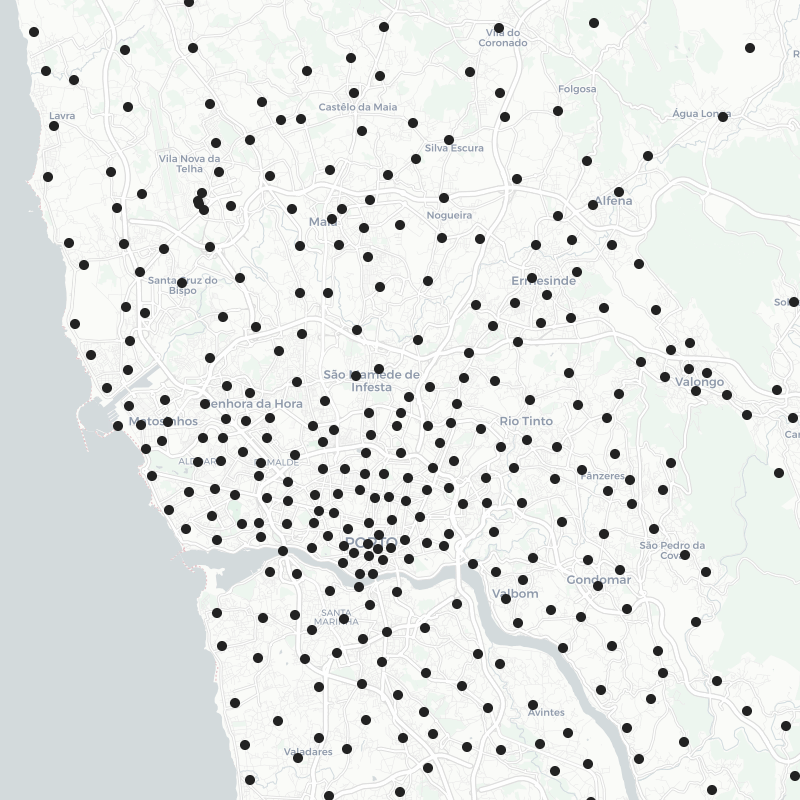}};
    
    \node[anchor=east, align=center, rectangle, draw=black] (city_gwr) at (3,-3.5) {City-level GWR \\ \includegraphics[height=2.5cm]{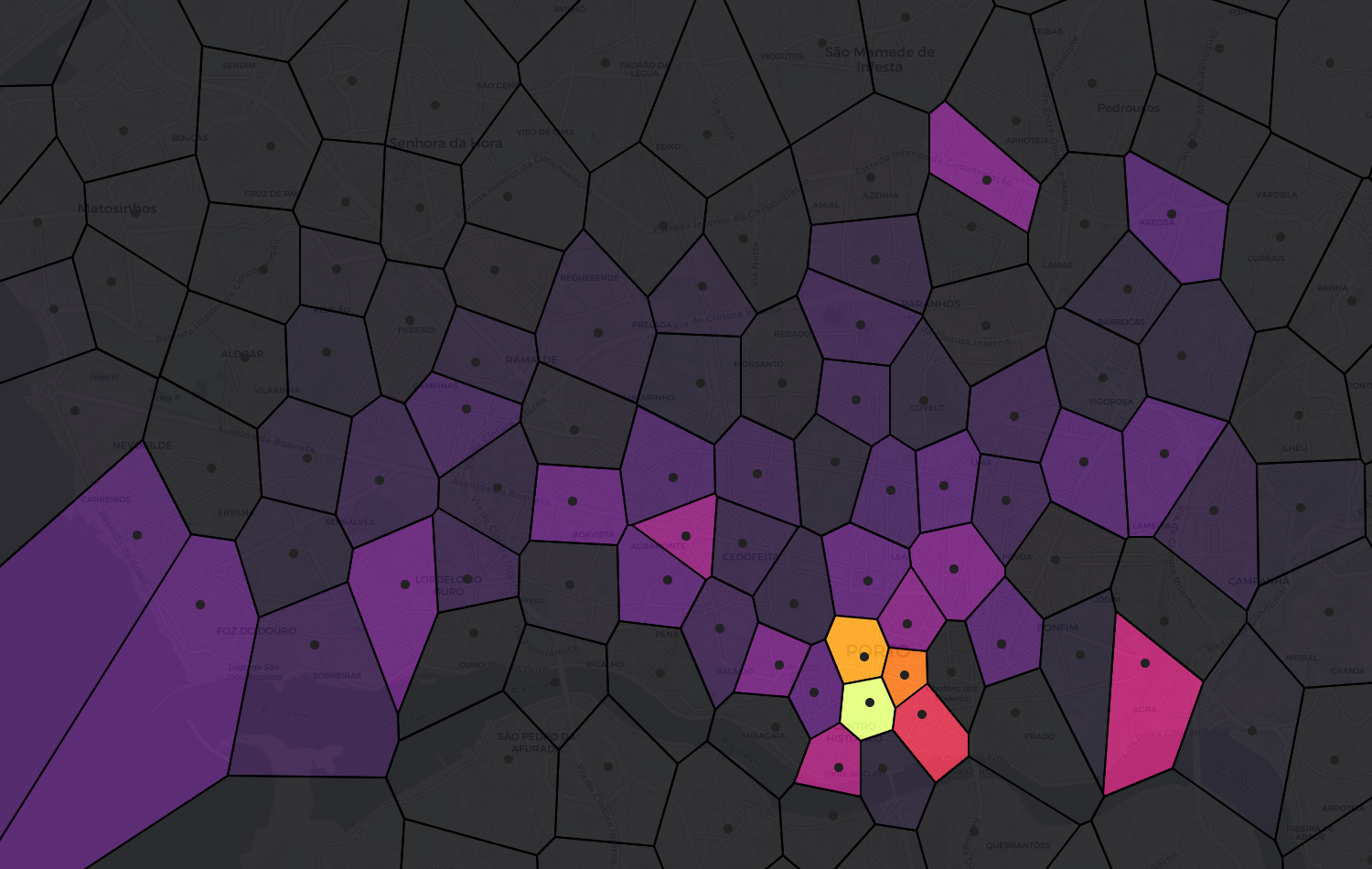}};

    \node[anchor=north west, align=center, rectangle, draw=black] (anomaly) at ($(input.north -| city_gwr.south) + (3.5,0)$) {Anomalies with geospatial interpretation \\ \includegraphics[height=6cm]{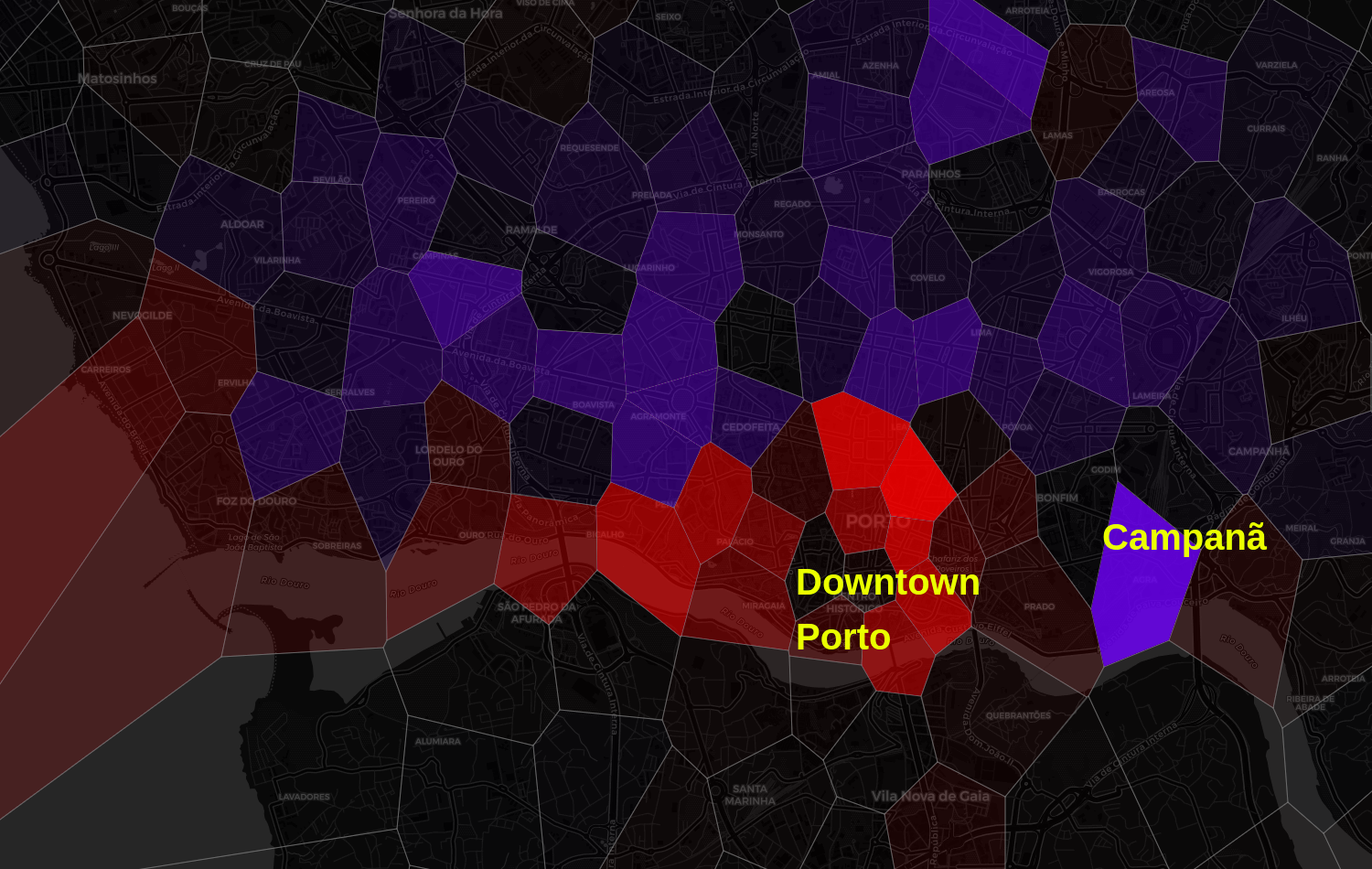}};
    
    \draw[-{Latex[scale=2]}] (region_gwr) -- (city_gwr);
    
    \draw[-{Latex[scale=2]}]  ($(input.south -| region_gwr.north) + (.5,0)$) -- ($(region_gwr.north) + (.5,0)$);
    \draw[-{Latex[scale=2]}]  ($(input.south -| city_gwr.north) - (1,0)$) -- ($(city_gwr.north) - (1,0)$);
    
    \draw[-{Latex[scale=2]}] (city_gwr) -- (anomaly);
    
    \node[draw, dotted,inner sep=2mm,label=below:Continual/lifelong learning,fit=(input) (region_gwr) (city_gwr)] {};
    
    \node[inner sep=2mm,label=below:Online inference,fit=(anomaly)] {};
    
\end{tikzpicture}
}
  \caption{Grow-when-required (GWR) networks learn online from unsupervised data streams without catastrophic forgetting. Our system mines city-wide patterns of life and detects unique cultural events as they occur in a mobility data stream.%
  }
  \Description{T}
  \label{fig:teaser}
\end{teaserfigure}
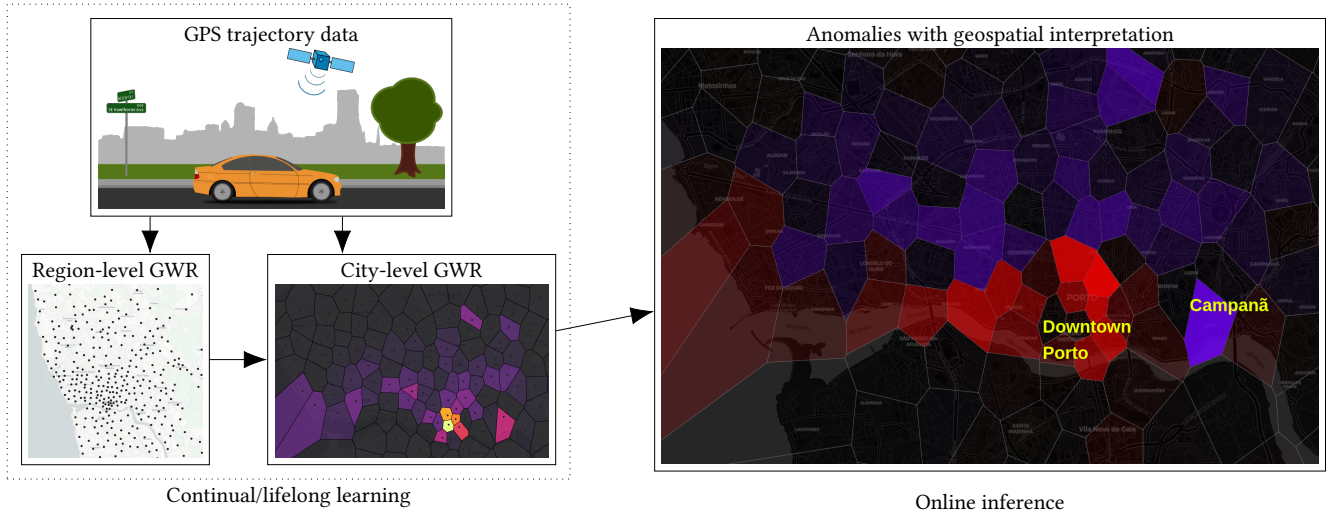

\maketitle

\section{Introduction}

What can we learn from observing the mobility of human populations? The study of human movement and its properties, causes, motivations, and constraints has a long and storied history, dating to at least the 19th century (see \cite{barbosa2018human} for a review). In its most modern form, the field increasingly uses a proliferation of positioning systems, including GPS, cell towers, and WiFi, to mine individual- and population-level patterns-of-interest \cite{toch2019analyzing}.  Geospatial professionals such as city planners, transportation managers, and ride-sharing developers have a compelling need to understand patterns of normal behavior and detect deviations from them in real time.

However, incrementally learning patterns from data streams poses unique challenges for machine learning (ML) systems. Over time, any number of factors may alter the statistical distribution of the data stream.  Some neighborhoods grow, and others decline; new businesses open, and old ones close their doors; a tourism boom may boost traffic, or an unexpected pandemic may shrink it rapidly. These \emph{concept drifts} are a significant danger to ML systems. Models pre-trained on historical data are unable to adapt to the altered patterns, but conversely, ML systems that na{\"i}vely train on new data will instead forget older information which may be useful in the future. Since each gradient update optimizes only for the current mini-batch, and not previous historical data, any learned knowledge is rapidly forgotten in subsequent weight updates. This problem is commonly known as \emph{catastrophic forgetting}.

A second issue for many ML methods, especially in deep learning, is the difficulty of explaining or interpreting their internal reasoning. As a result, they are often described as ``black box'' models. For anomaly detection, this is particularly problematic: with neither ground-truth labels nor interpretable reasoning, end users have little basis to trust the model's decisions. Taken together, these distinct challenges combine into a much greater problem.  We know that any ``black box'' system will encounter concept drifts, but because we cannot inspect the model's reasoning or compare its predictions with ground truth, we cannot easily assess these drifts and their harmful effects on the model.

Instead, we propose an approach which borrows a modeling framework from computational biology and neurorobotics, two disciplines which have long researched how real-life organisms operate under dynamically changing environments and applied these insights to artificial systems. The proposed approach uses Grow When Required (GWR) networks of bio-inspired episodic memories, which are explicitly designed to acquire unsupervised knowledge over time and protect it from subsequent degradation. We ensure that all encoded knowledge is accessible throughout and after the learning process: a set of normalcy patterns, or \emph{patterns of life}, enabling the model to detect and explain any deviations from typical aggregate mobility behavior. As an example, the proposed system detects multiple culturally important events in the classic Porto taxi dataset which, to our knowledge, have not been reported or modeled in previous work. We envision that these patterns-of-life models will be very useful in continually mining unsupervised GPS streams, due to their natural robustness to concept drifts, especially in domains where human-machine trust is paramount. %

\section{Related work}

\subsection{Machine learning and human mobility}
The growth of large-scale public mobility datasets has encouraged a parallel development of machine-learning methods to generate predictions and mine patterns of interest from data (as reviewed by \citet{toch2019analyzing}). In the unsupervised setting most relevant to this work, researchers often use clustering methods, especially $k$-means \cite{ashbrook2003using, cao2010mining, andrienko2012visual, reades2007cellular}, and density-based \cite{nanni2006time, cao2010mining, lee2007trajectory, pelekis2009clustering, andrienko2011movement} clustering.  Hierarchical \cite{ying2014mining, lin2013detecting} and spectral clustering \cite{rosler2013using} may also be used to group related users, places, or trajectories, and generative topic models such as latent Dirichlet allocation may reveal underlying routines or lifestyles \cite{farrahi2011discovering, zion2017learning, deb2015discovering}. Some recent works have incorporated variational autoencoders \cite{gao2019predicting}, graph convolutional networks \cite{jenkins2019unsupervised}, and spectral graph wavelets from manifold learning \cite{watson2020identifying}.

Recently many authors have used taxicab data-sets for targeted prediction problems, such as prediction of GPS points within the trajectory (including the final destination and/or arbitrary points) \cite{de2015artificial, zhang2018multi, rossi2019modelling, zhang2019prediction, li2020fast, ebel2020destination, liao2021taxi}, passenger demand \cite{moreira2013predicting, saadallah2018bright, le2019neighborhood, rodrigues2020spatiotemporal}, and travel time \cite{hoch2015ensemble, gupta2018taxi, fu2019deepist, lan2019travel, das2019map, abbar2020stad}. A smaller community of authors has also used these data-sets to develop unsupervised methods which aim to detect unusual or anomalous trajectories \cite{lam2016concise, wu2017fast, keane2017detecting, irvine2018normalcy, song2018anomalous, liu2020online}. \citet{lam2016concise} employs a nearest-neighbor search to determine anomalous driving behavior; \citet{irvine2018normalcy} compute a probability distribution over learned sequences from tokenized trajectories; \citet{wu2017fast} use a reinforcement model of driver preferences and behavior. Other authors train recurrent neural networks in both supervised and unsupervised settings. \citet{song2018anomalous} label a set of trajectories as anomalous and train a model to classify them, while \citet{liu2020online} propose a deep recurrent variational autoencoder.

\subsection{Approaches to lifelong learning}

The problem of catastrophic forgetting in neural networks was first reported more than 30 years ago \cite{mccloskey1989catastrophic, ratcliff1990connectionist} and has been a highly active research domain ever since.  The resulting literature of \emph{lifelong learning} (a.k.a. continual learning, incremental learning) has largely crystallized into three distinct approaches (see  \cite{parisi2019continual} for a thorough review). First, replay (a.k.a. rehearsal) simply stores a small set of training inputs (such as, but not always, an online random sample) and periodically retrains the model on it \cite{isele2018selective, riemer2019scalable, chaudhry2019tiny, kim2020imbalanced}. Second, more complex regularization approaches such as \cite{li2017learning, kirkpatrick2017overcoming, zenke2017continual} all add additional loss functions to encourage the model to preserve some previous knowledge while training on new data. Finally, and most relevant to our work, \emph{dynamic architectures} change structural properties of the network itself to accomodate the incremental accumulation of new knowledge. Neurons can be inserted or removed; their connections, rewired into entirely new synaptic organizations; entire sets of weights, ``frozen'' or trained depending on the input stream. The scale of changes could be as small as a single neuronal unit or can expand to mutate large sub-networks of the model, e.g., \cite{zhou2012online, xiao2014error, rusu2016progressive, draelos2017neurogenesis, yoon2017lifelong}.

One of the most well-studied families of dynamic neural architectures is the self-organizing map (SOM), developed by Kohonen in the early 1980s (e.g., \cite{kohonen1982self}). SOMs consist a set of neurons, or units, undergoing competitive learning: each training example is matched to a unit with the most similar features (the \emph{best-matching unit}, or BMU), and used to train the BMU and other nearby units within a predefined neighborhood function. A later, but similar, alternative is the neural gas, which introduced connections between related neurons \cite{martinetz1991neural}, i.e., the famous Hebbian learning principle that ``cells that fire together, wire together.'' Neither of these formulations allows for \emph{neurogenesis}, or the creation of new neurons. However, they inspired a vast array of new ``growing'' networks, many explicitly intended for lifelong learning applications \cite{bauer1997growing, fritzke1995growing, prudent2005incremental, marsland2002self}. Most importantly for this work, the grow when required (GWR) network \cite{marsland2002self} introduced a biologically-inspired \emph{habituation} mechanism for a network to adjust its size in response to inputs.

These networks have been especially well-studied for autonomous agents and neurorobotics \cite{jun2003robot, neto2007visual, parisi2018lifelong, ezenkwu2019unsupervised, contreras2019vision, pitonakova2020robustness, duczek2021continual}, including by the original authors of GWR \cite{marsland2005line}, because these kinds of networks coincide with a key intrinsic motivation of any autonomous system: to progressively learn to ignore sensory inputs which are already well-known, and to identify anything else \cite{nehmzow2013novelty}. This principle is not unique to robotics, also finding applications as diverse as retail sales \cite{decker2005market}, action and emotion recognition \cite{parisi2017lifelong, mici2018self, barros2019personalized}, video surveillance \cite{acevedo2011clustering, sun2017online}, medical imaging \cite{angelopoulou2005automatic, de2009detection, netto2012automatic}, and genomics \cite{todorov2020tinga}. However, few authors have applied any SOM or related method to transportation applications (rare exceptions include \cite{rudloff2010detecting, lopez2017spatiotemporal, beura2018defining}), and to our knowledge, we are the first to apply either GWR or a lifelong-learning SOM to this domain.

\section{Problem Description}

Let $\mathcal{T} = T_1, T_2, \ldots$ be the data stream of all trajectories in the dataset, in order of occurrence, where a trajectory is a variable-length series of latitude-longitude points $T_i = (\phi_1, \lambda_1), \ldots, (\phi_n, \lambda_n)$. For computational convenience (to calculate distances in meters and kilometers simply), we transform these into UTM coordinates $T_i = (x_1, y_1), \ldots, (x_n, y_n)$.

We assume that for each trajectory $T_i$, there are some number of latent factors $\Psi$ which influence $T_i$, most notably the geospatial probability distribution of where $T_i$ originates. Examples might include city-wide factors such as tourist volume, weather, or academic schedules and geospatially-local factors such as new construction or road closures. The identity of all the factors $\Psi$, or even the number of factors $\vert \Psi \vert$, is itself impossible to observe and intractable to learn. However, we know that (some subset of) $\Psi$ may vary over time.  Therefore, even without observing $\Psi$ directly, we aim to detect \emph{changes} in $\Psi$ by observing changes in the trajectory stream $\mathcal{T}$.

Our problem, then, is to continually monitor a GPS trajectory stream $\mathcal{T}$, here from the Porto taxi dataset, and learn to construct a model $f$ of $\mathcal{T}$'s normal behavior without supervision. When an interval of trajectories $T_i, \ldots, T_j \in \mathcal{T}$, taken in the aggregate, deviates from ordinary behavior, $f(T_i, \ldots, T_j)$ must detect the anomaly and provide enough spatiotemporal explanation of its own reasoning that a human observer can interrogate the purported anomaly (often called a ``white box'' model). We consider an anomaly detection correct if, and only if, both of the following conditions are met: (1) the model provides sufficient explanation of its reasoning that a human user can quickly search for sociocultural information to confirm or reject the detection, and (2) the detection is in fact accurate: additional pieces of evidence, such as contemporaneous sources, confirm a plausible explanation for unusual activity in the specified location and time.

The model $f$ can detect changes in the data stream $\mathcal{T}$; the human-model team can detect changes in the cultural factors $\Psi$. This is a notably more difficult problem than ordinary anomaly detection, which is typically only concerned with detection accuracy relative to a provided ground truth.  Instead, our problem formulation mandates explanation and interaction between $f$ and a human partner: \emph{we only count an anomaly detection as correct if it has a human-readable explanation.} However, we argue that in the setting of fully unsupervised learning, requiring human-in-the-loop explanation and validation is essential in order to establish trust in the final model.

\section{Methods}

\subsection{Interpretation of a trained GWR network}
One of the advantages of a GWR network, as opposed to a ``black box'' model such as a deep neural network, is that each component provides a convenient interpretation. The GWR can be viewed as an online clustering of the data stream.  Each neuron has a weight vector $\mathbf{w}_i$ which represents a prototype input (such as a GPS location or citywide traffic pattern), and the model seeks to find a limited set of prototypes which are adequate to match all observations. We will show in \cref{sec:modeling_dates,sec:discovering_new_events} that a properly-constructed set of prototypes $\{ \mathbf{w}_i \}$ can be viewed as a model of the trajectory stream $\mathcal{T}$'s normal behavior, with each $i$ representing a different sub-type of ``behavior.''  We refer to these prototypical behaviors, operationalized as GWR weight vectors, as the \emph{patterns of life} extracted from the data stream.  Unlike some techniques such as $k$-means or Gaussian mixtures, the number of prototypes need not be known in advance or tuned as a hyperparameter: it will be learned during model training.  

Each neuron $i$ also has a \emph{habituation}, a scalar value $\eta_i$ which can be viewed as an inverse of the probability mass or support of each prototype.  It is possible to recover from $\eta_i$ an approximate measure of how frequently neuron $i$ was activated, that is, the proportion of observations with which neuron $i$ is associated. %

The GWR can also be viewed as a one-layer autoencoder, a type of (typically deep) neural network which accepts an input $\mathbf{x}$ and attempts to learn a function to reconstruct it, $f: \mathbf{x} \rightarrow \mathbf{x}$ given information constraints. These autoencoders are frequently used for unsupervised anomaly detection, where a sudden increase in reconstruction error implies an input $\mathbf{x}$ which does not match the model's training set.  Note that the error $d_b$ described below is, essentially, a $\ell_2$-based reconstruction loss (\cref{eq:activity}), and the update formula for $\mathbf{w}_i$ (\cref{eq:w_update}) a gradient descent step to minimize this loss.  As in many unsupervised anomaly detections, an increase in the reconstruction loss $d_b$ (or equivalently, drop in $a$) implies an anomaly, but instead of a black-box result, the input $\mathbf{x}$ can be compared element-wise with the most relevant prototype, $\mathbf{w}_b$ (see \cref{fig:known_events,fig:discovered_anomalies}).

\subsection{Detailed training procedure}

We describe the GWR algorithm formally in this section. For hyperparameter values and a graphical depiction of the algorithm, please refer to the Appendix.

\subsubsection{Initialization}
\label{gwr_initialization}
We begin with a set of two neurons represented by their weight vectors, $A = \{ \mathbf{w}_1, \mathbf{w}_2 \}$.   We also require a representation of whether two neurons are connected; we use a $2 \times 2$ (sparse) adjacency matrix $\mathbf{E} = \mathbf{0}$. Note that this graph is undirected and $\mathbf{E}$ must remain symmetric. \emph{For brevity, we will omit duplicate assignments to $\mathbf{E}$, but any equality or assignment $\mathbf{E}_{ij} = k$ also implies $\mathbf{E}_{ji} = k$.} We will use the notation $\mathcal{N}(i)$ to refer to the neighbors of neuron $i$, that is, the set $\{j \mid \mathbf{E}_{ij} > 0 \}$.

\subsubsection{Finding matching units}
\label{gwr_matching}
We begin iterating over each input $\mathbf{x}$ in the data stream. We find the best matching unit (BMU) $b$, the neuron $\mathbf{w}_i$ with the least distance to the input $\mathbf{x}$. We also find the second-best matching unit $s$ similarly, by finding the BMU of all neurons except $b$,
\begin{align}
    b &= \arg \min_{i \in A} \lVert \mathbf{x} - \mathbf{w}_i \rVert ^ 2 & 
    s &= \arg \min_{i \in A \setminus b} \lVert \mathbf{x} - \mathbf{w}_i \rVert ^ 2
\end{align}
A connection between $b$ and $s$ is made, if it does not already exist, by setting $\mathbf{E}_{sb}=1$. This forms a synapse between the best two neurons for the given input, analogous to Hebbian learning principles from neurobiology \cite{martinetz1991neural}.

\subsubsection{Network activity}
\label{gwr_activity}
We calculate the \emph{activity} of the network, $a$, as a simple transformation of the network's reconstruction error $d_b$,
\begin{align}
    d_b &= \lVert \mathbf{x} - \mathbf{w}_b \rVert ^ 2 &
    a &= \exp(-d_b) \label{eq:activity}
\end{align}
Note that this provides a performance metric $a \in (0, 1]$ which the network seeks to optimize. A perfect reconstruction results in $a=1$, whereas $\lim_{d_b \rightarrow \infty} a = 0$.

\subsubsection{Neurogenesis}
\label{gwr_neurogenesis}
More importantly, however, the activity $a$ provides us a criterion to control the creation of new neurons. Note that $a$ represents the performance of the \emph{best} matching unit: by definition, all the other neurons perform worse for this particular input $\mathbf{x}$.  Therefore, if $a$ is low, then \emph{none} of the units matched the input well. 

If $a$ is below an activity threshold $a_T$, we add a new neuron $r$,
\begin{align}
    & \mathbf{w}_r = (\mathbf{x} + \mathbf{w}_b) / 2 & A \leftarrow A \cup \mathbf{w}_r \\
    & \mathbf{E}_{rb} = \mathbf{E}_{rs} = 1 & \mathbf{E}_{sb} = 0
\end{align}
In other words, the weights of the new neuron $\mathbf{w}_r$ are the average of the input $\mathbf{x}$ and the best-matching unit $\mathbf{w}_b$. In the adjacency matrix $\mathbf{E}$, $r$ is inserted between the two neurons with the best reconstruction $b$ and $s$, so that it is connected to each of them, but they no longer connect to each other.
\subsubsection{Neuron update}
\label{gwr_update}
If, and only if, the activity threshold above is \emph{not} reached, we instead update the weights for the BMU $b$ and any other neuron attached to it $\mathcal{N}(b)$. A simple update rule would be,
\begin{align*}
    \forall i &\in b \cup \mathcal{N}(b),& \Delta \mathbf{w}_i &= \epsilon_i(\mathbf{x} - \mathbf{w}_i)
\end{align*}
where $\epsilon_i$ is a learning rate for neuron $i$. In practice, it is typical to use a relatively large learning rate for the BMU, $\epsilon_b$, and a smaller learning rate $\epsilon_n$ for its neighbors.
\subsubsection{Final steps}
\label{gwr_end}
The iteration concludes by altering $b$'s connections to neurons \emph{other than $s$} in the adjacency matrix $\mathbf{E}$. Until this point, we have only inserted zeros and ones into $\mathbf{E}$.  However, the elements of $\mathbf{E}$ actually represent synapse ``ages,'' that is, $\mathbf{E}_{ij}$ is the number of examples observed since $i$ and $j$ matched together as the two best neurons.  We increment the age of all $b$'s synapses, other than $s$. 
\begin{align}
    \forall i &\in \mathcal{N}(b) \setminus s, & \mathbf{E}_{ib} \leftarrow& \mathbf{E}_{ib} + 1 
\end{align}
Recall that we already reset the age of the $s$--$b$ synapse to $\mathbf{E}_{sb} = 1$ previously (\cref{gwr_matching}). If a synapse exceeds a certain age threshold $\mu_\mathrm{max}$ it may be safely pruned, and if a neuron $i$ loses all connections $\mathcal{N}(i) = \varnothing
$ then it may also be removed from the GWR network. This concludes the processing of the input $\mathbf{x}$, and we loop back to \cref{gwr_matching} and repeat for each input $\mathbf{x}$ in the data stream.

\subsubsection{Habituation}
\label{gwr_habituation}
The above GWR implementation is nearly complete, but one issue remains. Note that neurogenesis (\cref{gwr_neurogenesis}) and neuron updates (\cref{gwr_update}) are mutually exclusive. Since the network is initially untrained, the activity $a$ will likely be low and a new neuron will be created---precluding any training of existing neurons. For the next input, the existing neurons will remain poor, creating another new neuron---and preventing training. This loop of repeated neurogenesis leads to a network much larger then it needs to be; a smaller set of well-trained neurons would have been adequate.  After all, the GWR network should only grow \emph{when required}.

The solution is \emph{habituation}, a computational model of how synapses lose efficacy as they are repeatedly activated \cite{stanley1976computer, marsland2002self}. We modify the definition of a neuron $i$ to include both its weight vector $\mathbf{w}_i$ and a scalar \emph{firing counter}, which is initialized to $\eta_i = 1$. Crucially, $\eta_i$ decays each time $\mathbf{w}_i$ is trained, as shown below. Therefore, during training $\eta_i$ will give us a measure of whether neuron $i$ is old and well-trained, or newly-formed and in need of tuning.

Habituation provides the network with two desirable properties.  First, adding a new neuron requires not only that $a < a_T$ (see \cref{gwr_neurogenesis}), but also that the best-matching unit have $\eta_b < f_T$ for some firing threshold $f_T$. If $\eta_b$ is close to 1, then $\mathbf{w}_b$ has not yet been adequately trained, and the network will choose to improve its weights $\mathbf{w}_b$ rather than adding a new neuron.

Second, we modify the rules for updating neurons presented in \cref{gwr_update}:
\begin{align}
    \forall i &\in b \cup \mathcal{N}(b),& \Delta \eta_i &= \tau_i \kappa (1 - \eta_i) - \tau_i \label{eq:eta_update} \\
    && \Delta \mathbf{w}_i &= \epsilon_i \eta_i (\mathbf{x} - \mathbf{w}_i) \label{eq:w_update}
\end{align}
where $\kappa$ and $\tau_b, \tau_n$ are additional hyperparameters controlling the rate of habituation. Like the learning rates $\epsilon_b, \epsilon_n$, the rate $\tau$ is larger for the BMU $b$ than its neighbors $n$.

Note that $\eta_i$ decreases whenever $\mathbf{w}_i$ is updated (\cref{eq:eta_update}) and also controls the size of $\mathbf{w}_i$'s update (\cref{eq:w_update}). Therefore, \emph{the more neuron $i$ is trained, the more resistant it becomes to training.} This helps to prevent catastrophic forgetting when the data stream changes, by explicitly preventing the degradation of knowledge stored within well-trained neurons. 

\subsubsection{Summary}
Suppose example $\mathbf{x}$ is best-matched with BMU $b$. The network's performance is the activity $a$. There are three cases to consider:
\begin{enumerate}
    \item The activity $a$ is close to 1: The GWR network performs well. The weight $\mathbf{w}_b$ will be updated, but the magnitude of the term $\mathbf{x} - \mathbf{w}_b$ is small. As a result, $\mathbf{w}_b$ does not change much (\cref{eq:w_update}).
    \item The activity $a$ is below the threshold $a_T$, but the firing counter $\eta_b$ is not yet below its threshold $f_T$: The best-matching unit performs poorly, but its weights are based on relatively little training data. Therefore, we train the weights $\mathbf{w}_b$. The magnitude of $\mathbf{x} - \mathbf{w}_b$ (\cref{eq:w_update}) is large, and $\mathbf{w}_b$ will change more significantly.
    \item The activity $a$ is below the threshold $a_T$, and the firing counter $\eta_b$ is below threshold $f_T$: The best-matching unit performs poorly, but its weights have been trained previously on large amounts of data. Rather than degrade that pre-existing knowledge with a weight update (causing catastrophic forgetting), we preserve it and insert a new neuron $r$ instead.
\end{enumerate}

\begin{figure}
    \centering
    \begin{subfigure}{0.15\textwidth}
        \includegraphics[width=\textwidth]{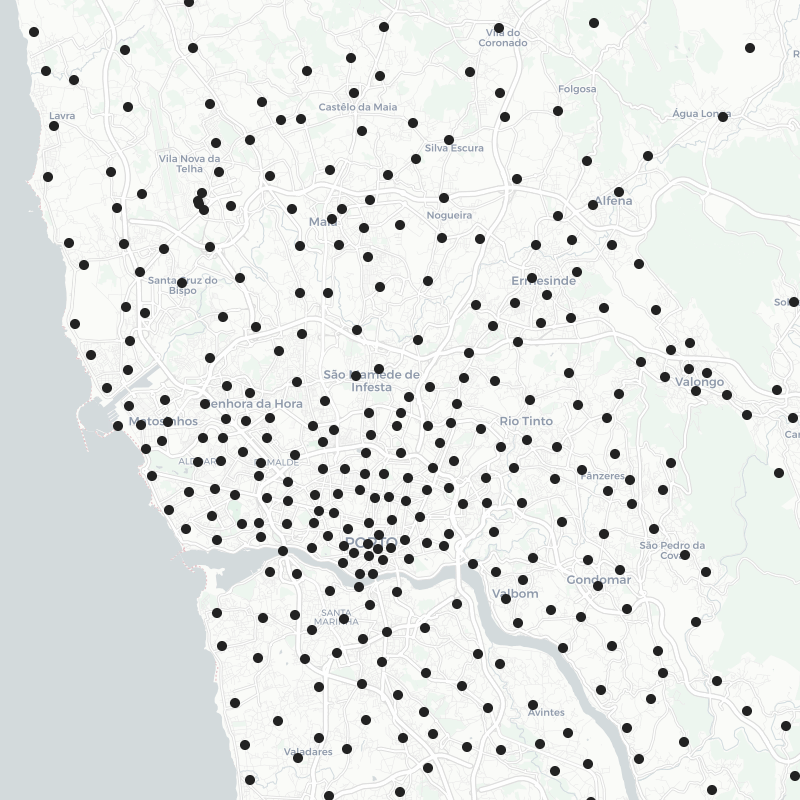}
        
        \caption{\label{fig:gwr_origins:a}}
    \end{subfigure}
    \hfil
    \begin{subfigure}{0.15\textwidth}
        \includegraphics[width=\textwidth]{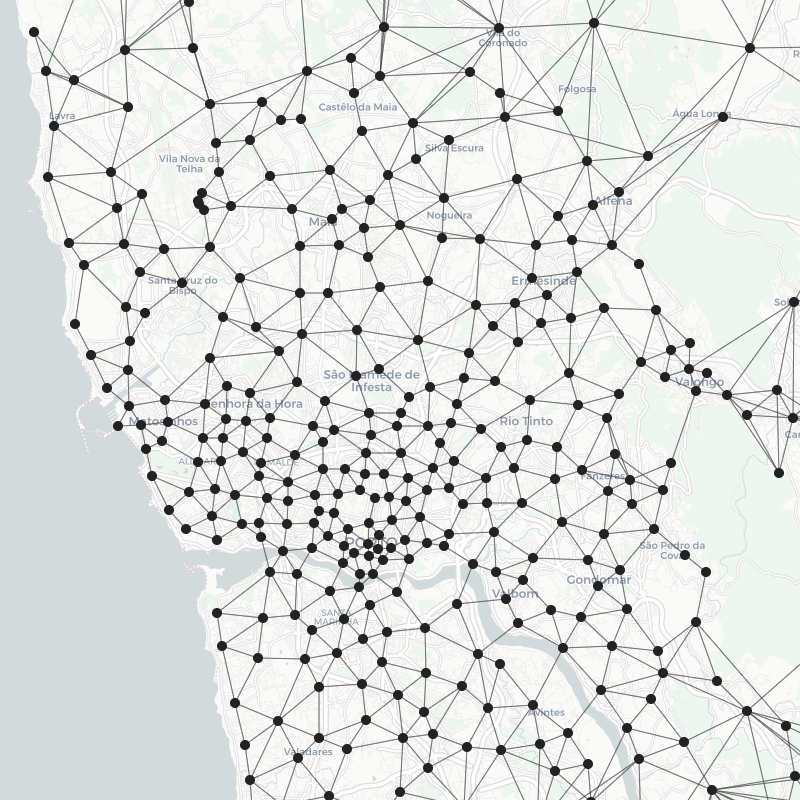}
        \caption{\label{fig:gwr_origins:b}}
    \end{subfigure}
    \hfil
    \begin{subfigure}{0.15\textwidth}
        \includegraphics[width=\textwidth]{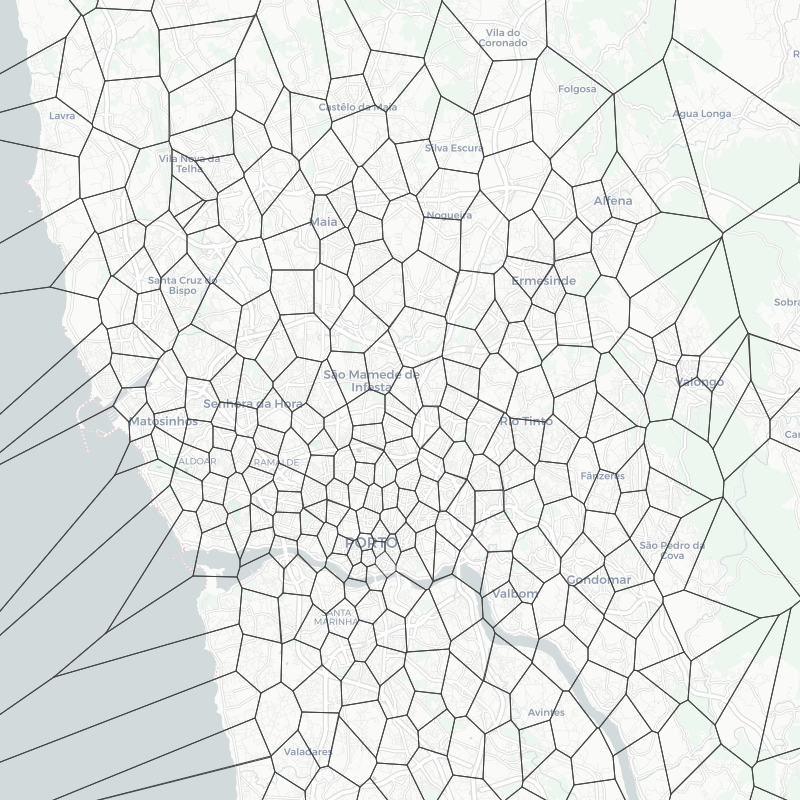}
        \caption{\label{fig:gwr_origins:c}}
    \end{subfigure}
    \caption{GWR of taxi origins. \textbf{\protect\subref{fig:gwr_origins:a}}: the weight vector $\mathbf{w}_i$ of each neuron $i$. Note that regions with high traffic density (e.g., downtown) have a higher density of neurons than peripheral areas due to habituation. \textbf{\protect\subref{fig:gwr_origins:b}}: Neurons with their synapses $\mathbf{E}$, which connect similar neurons. \textbf{\protect\subref{fig:gwr_origins:c}}: Receptive fields of each neuron produced via a Voronoi decomposition. Each region $R_i$ is a convex polygon containing all points $\mathbf{x}$ which would activate neuron $i$, i.e., all points for which $i$ would be the best-matching unit (BMU). Note that regions with higher neuron density have smaller receptive fields, encoding finer spatial detail.}
    \label{fig:gwr_origins}
\end{figure}

\section{Results}

Taxicabs are a particularly rich source of publicly-accessible research data, including from Porto \cite{moreira2013predicting}, San Francisco \cite{epfl-mobility-20090224}, New York City \cite{nyctaxi}, Rome \cite{roma-taxi-20140717}, and Beijing \cite{yuan2010t-drive}. Of these, the Porto dataset has been especially thoroughly researched since its 2015 release for a ECML/PKDD machine learning competition. Here, we present the results of our model using the Porto dataset. 

\subsection{Partitioning Porto with GWR}

Our approach uses a hierarchical GWR with two levels.  In the first level, we spatially partition the Porto area into sub-regions learned alongside the second-level anomaly detector. Let $\mathcal{T} = T_1, T_2, \ldots$ be the stream of all trajectories in the dataset, in order of occurrence.  Let $\mathbf{o}_t$ be the originating point of the trajectory at time $t$, that is, the point at which the driver picks up the passenger.  We input each $\mathbf{x} := \mathbf{o}_t$ into a GWR where each neuron learns a two-dimensional weight vector and iterate over all trajectories in the dataset. The hyperparameters for this GWR can be found in the Appendix.

We display the results of this origin-point GWR in \cref{fig:gwr_origins:a}. On the left, we depict only the prototype origin points $\mathbf{w}_i$.  Note that regions with high traffic density (e.g., downtown) have a high density of neurons compared to suburban and rural areas. This illustrates the effects of habituation (\cref{eq:eta_update}) on neurogenesis; the GWR is only allowed to create new neurons when $\eta_i < f_T$, which occurs in regions with high data density where neurons are frequently activated.  Note also that there seems to be a minimum distance between neurons even in high-density regions.  This is a consequence of the activity threshold: new neurons can only be created if $a < a_T$.  Because of the relationship between $d_b$ and $a$ (\cref{eq:activity}), there is a one-to-one mapping between an activity threshold $a < a_T$, such that neurogenesis only occurs for low $a$, and a distance threshold $d_T$ such that neurogenesis only occurs when the best neuron's error $d_b > d_T$.  This allows us to control the distance between neurons; for example, to only create neurons when an input point is at least $d_T = 1$ km from all the network's neurons, we set $a_T = \exp(-1)$.  In \cref{fig:gwr_origins:b}, we further display the synapses $\mathbf{E}$. Recall that since synapses are added between two neurons $b$ and $s$ which are the closest to some point $\mathbf{x}$; thus, $\mathbf{E}$ tends to connect neurons which are spatially proximate.

\begin{figure*}
    \centering
    \begin{subfigure}{0.32\textwidth}
        \includegraphics[width=\textwidth]{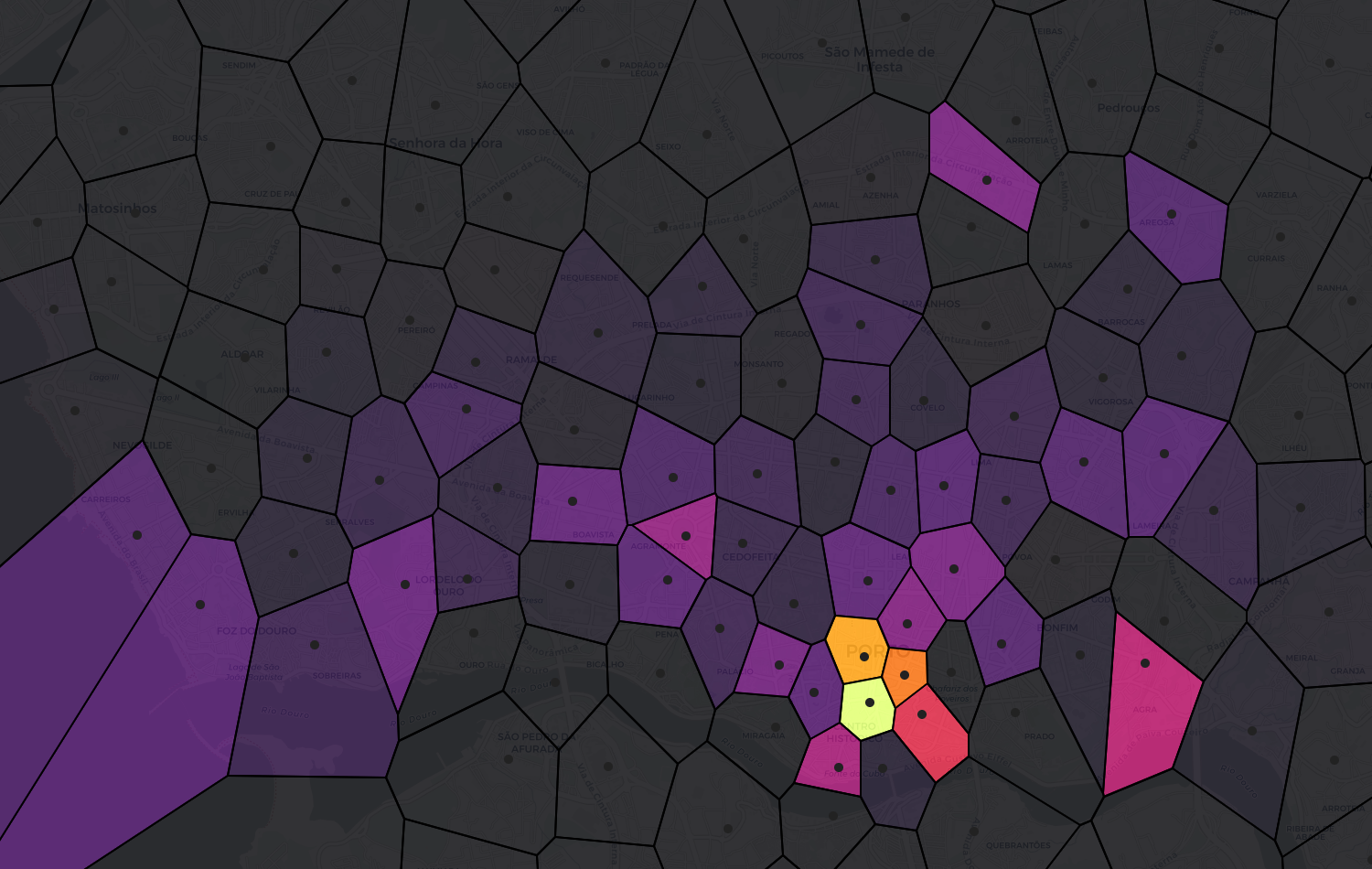}
        \caption{\label{fig:citywide_patterns:a}}
    \end{subfigure}
    \hfil
    \begin{subfigure}{0.32\textwidth}
        \includegraphics[width=\textwidth]{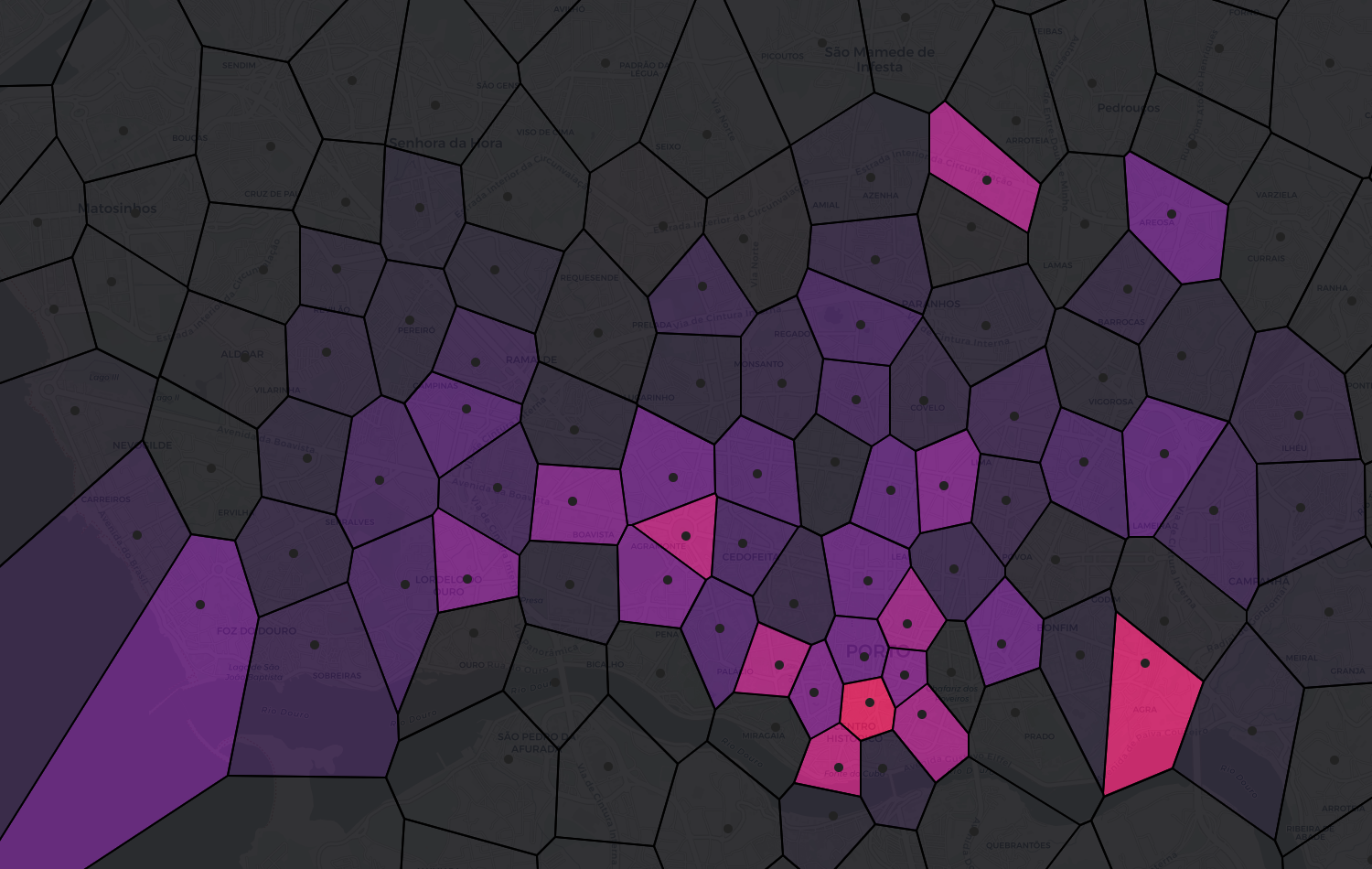}
        \caption{\label{fig:citywide_patterns:b}}
    \end{subfigure}
    \hfil
    \begin{subfigure}{0.32\textwidth}
        \includegraphics[width=\textwidth]{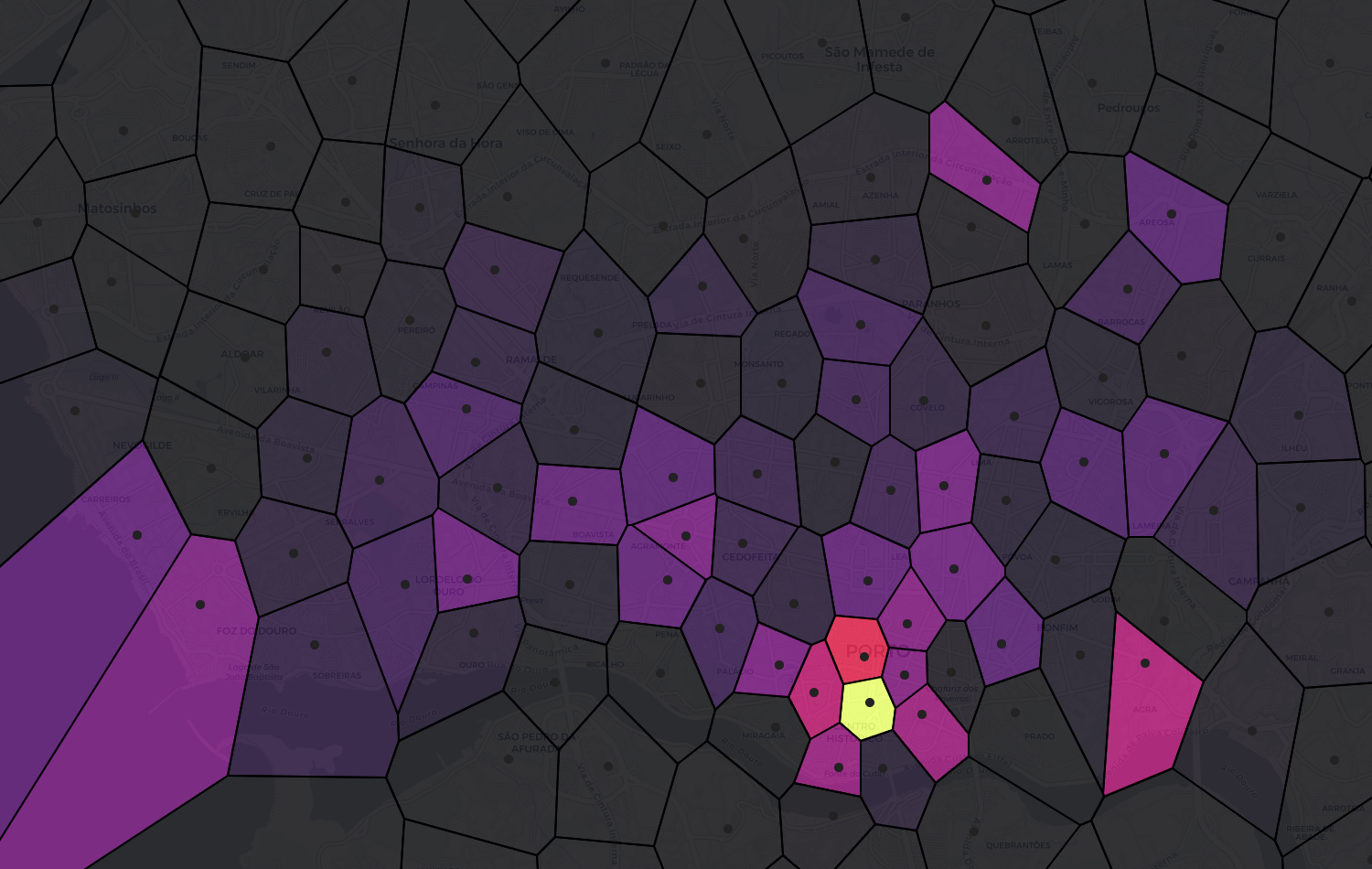}
        \caption{\label{fig:citywide_patterns:c}}
    \end{subfigure}
    \hfil
    \begin{subfigure}{0.32\textwidth}
        \includegraphics[width=\textwidth]{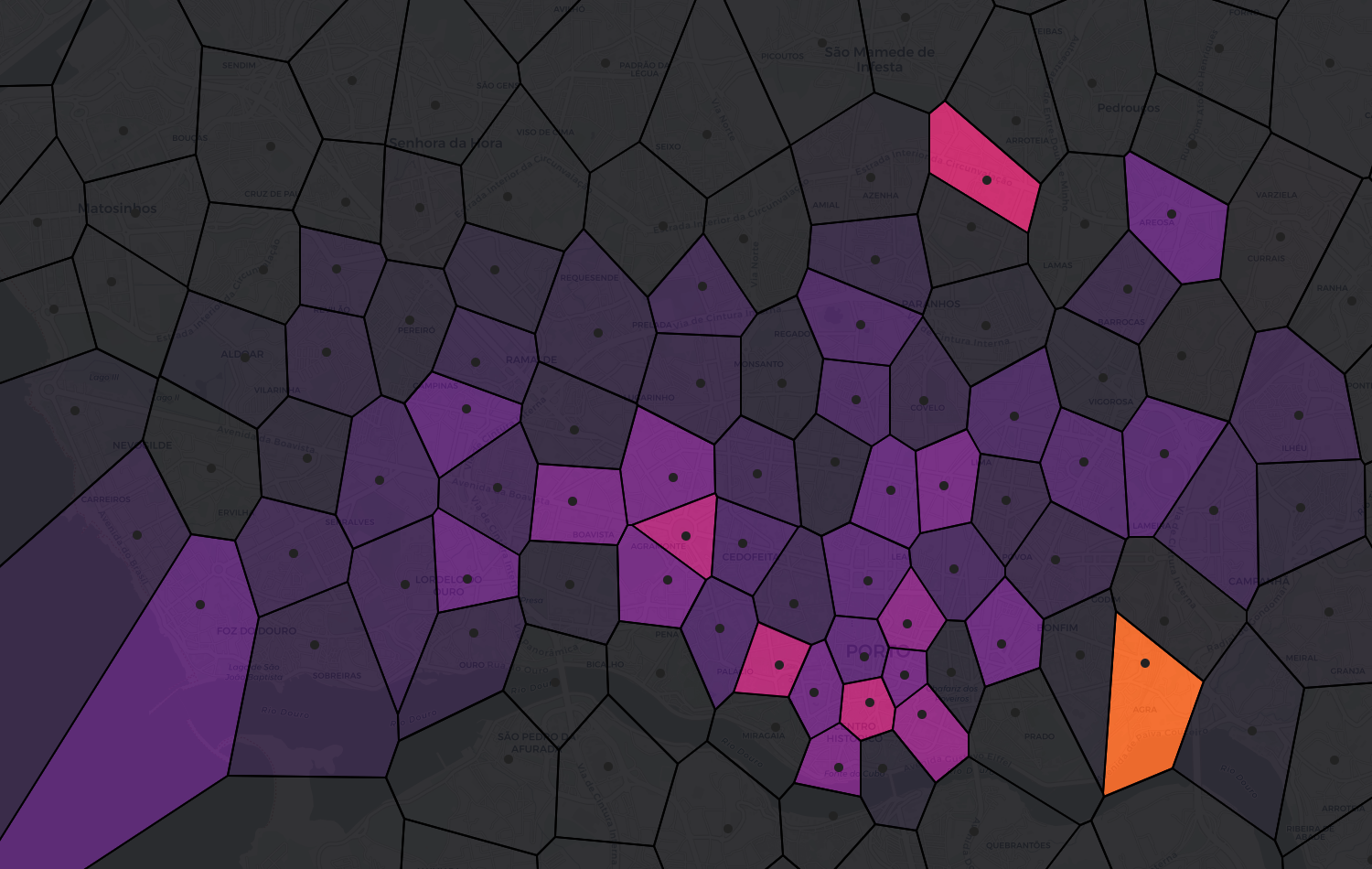}
        \caption{\label{fig:citywide_patterns:d}}
    \end{subfigure}
    \hfil
    \begin{subfigure}{0.32\textwidth}
        \includegraphics[width=\textwidth]{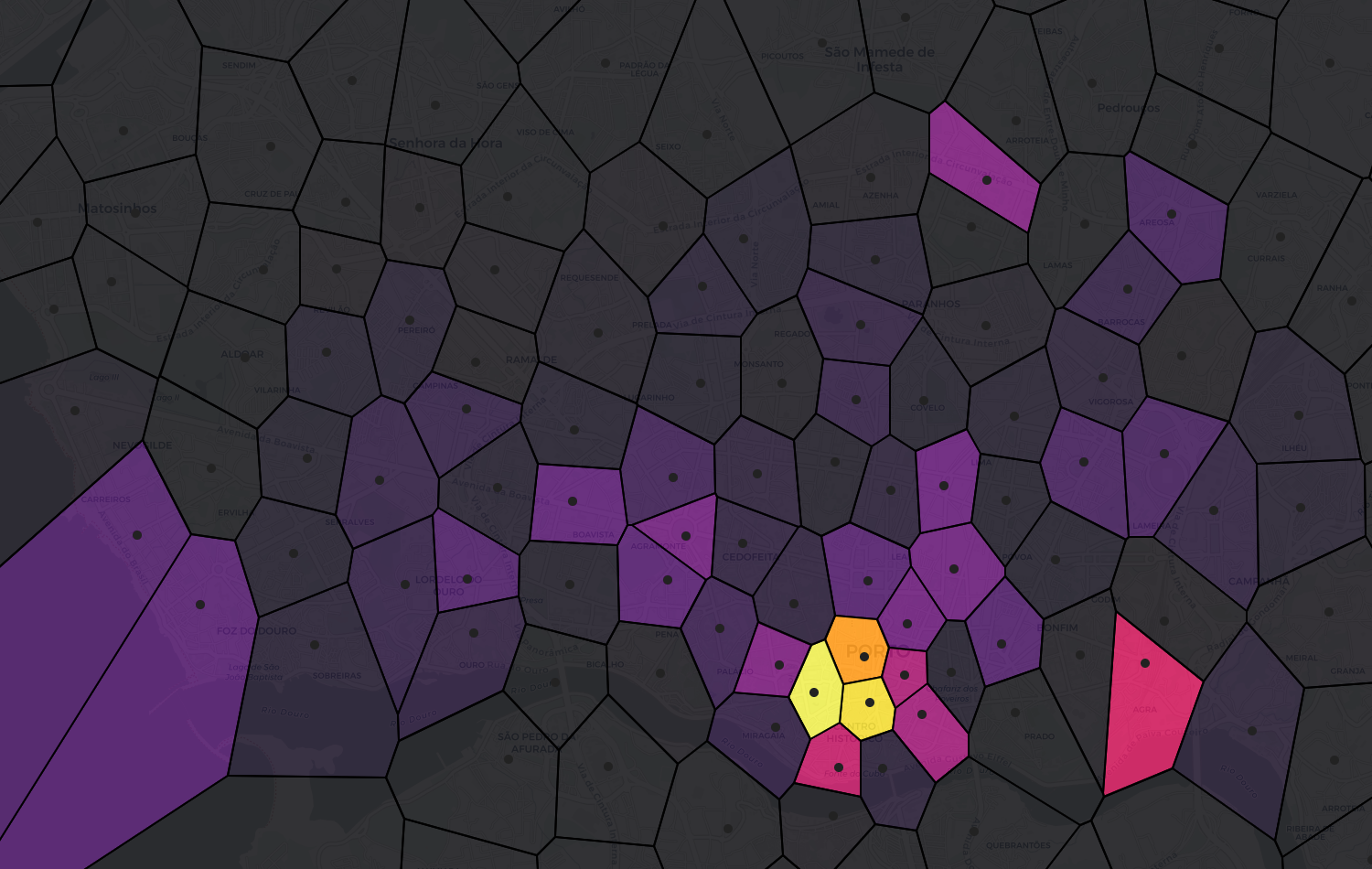}
        \caption{\label{fig:citywide_patterns:e}}
    \end{subfigure}
    \hfil
    \begin{subfigure}{0.32\textwidth}
        \includegraphics[width=\textwidth]{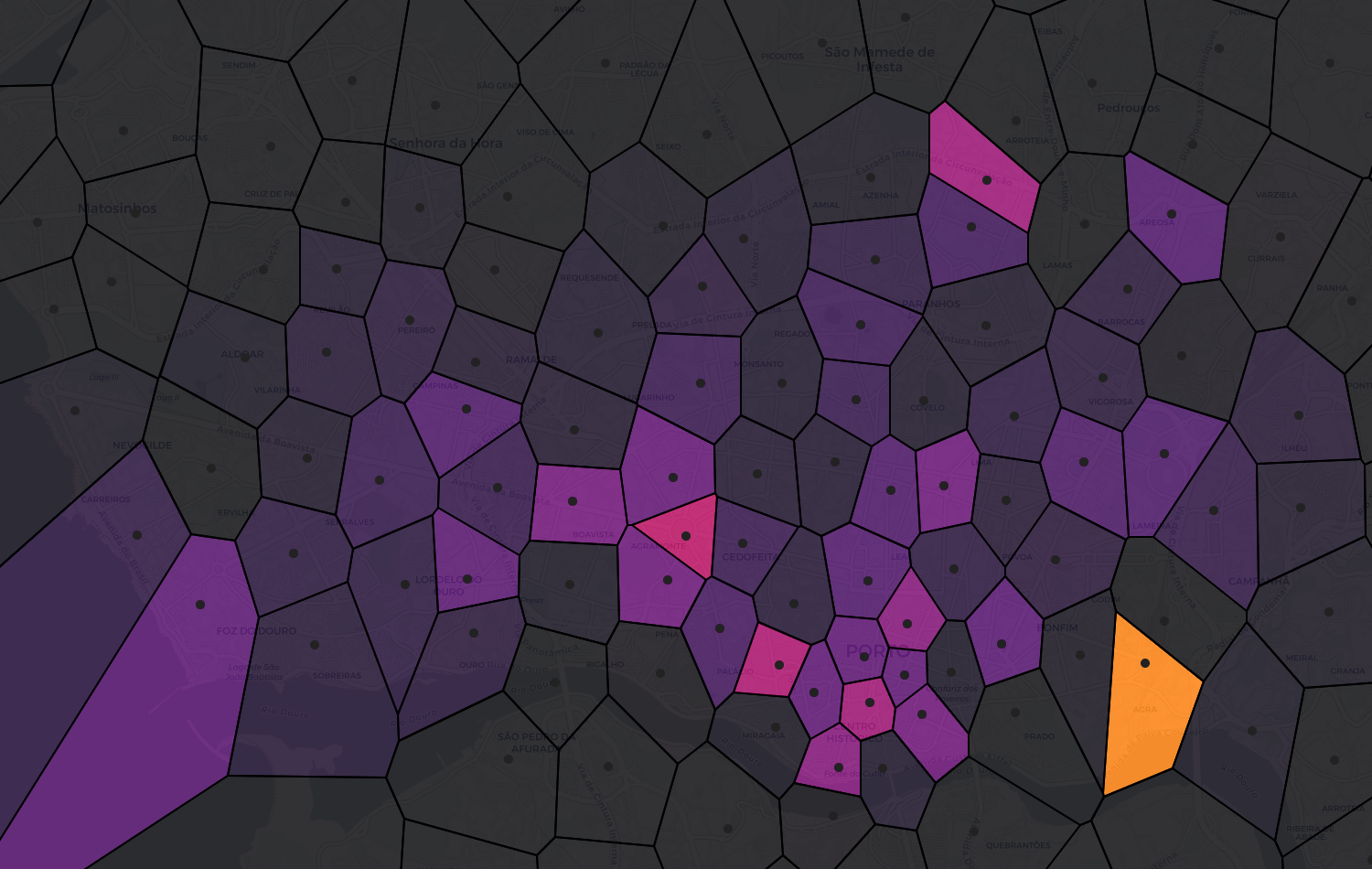}
        \caption{\label{fig:citywide_patterns:f}}
    \end{subfigure}
    \hfil

    \caption{Selected city-wide prototypes, constructed incrementally by a GWR model. The color of each cell indicates the density of taxi trajectories expected to originate in that region. On any given day, the city's traffic pattern is compared to the full set of prototypes; if none of them are a good match (i.e., $a'$ is low), the model determines that an anomaly occurred.}
    \label{fig:citywide_patterns}
\end{figure*}

\begin{figure*}
    \centering
    \includegraphics[width=.975\textwidth]{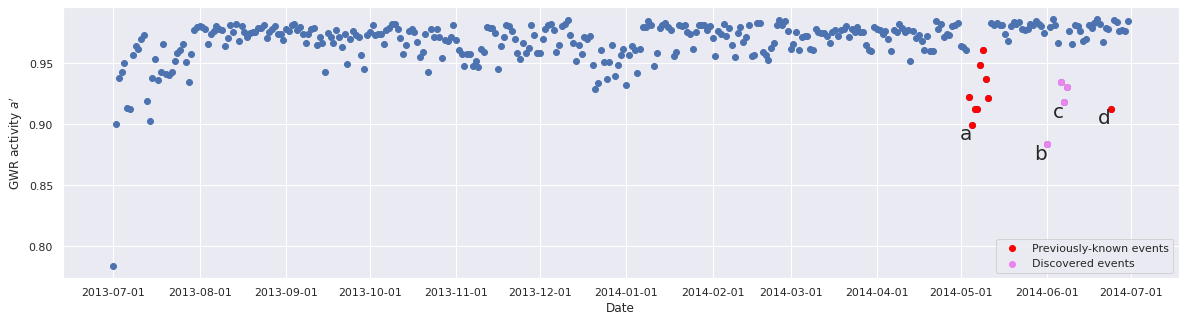}
    \caption{Activity score $a'$ for the second-level, city-wide GWR, with one value per day. The network takes roughly one month of data to reach a steady, high $a'$ value (e.g., August 2013); after that point, any sudden drop in $a'$ indicates the presence of an unusual event in taxi behavior. We confirm that culturally-well-known events such as \emph{Queima das Fitas} and \emph{Festa de S{\~a}o Jo{\~a}o do Porto} appear clearly as anomalies (red; see \cref{sec:modeling_dates}) but also uncover new anomalies not previously discussed in research on the Porto dataset (\cref{sec:discovering_new_events}). \textbf{(a)}: \emph{Queima das Fitas}; \textbf{(b)}: \emph{Serralves em Festa}; \textbf{(c)}: NOS Primavera Sound; \textbf{(d)}: \emph{Festa de S{\~a}o Jo{\~a}o do Porto}.}
    \label{fig:activity_plot}
\end{figure*}

\begin{figure*}
    \centering
    \begin{subfigure}{0.32\textwidth}
        \includegraphics[width=\textwidth]{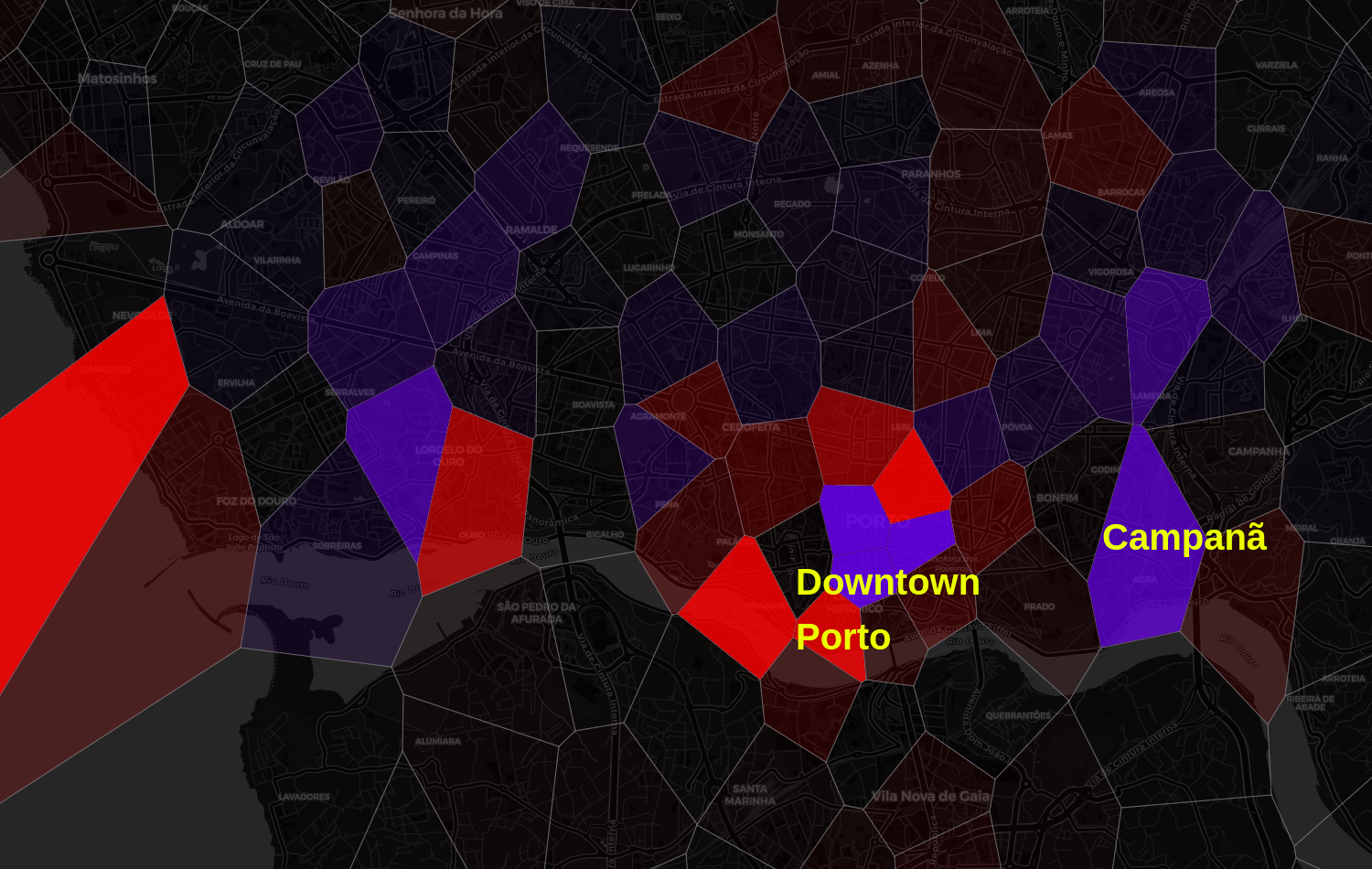}
        \caption{\label{fig:known_events:a}}
    \end{subfigure}
    \hfil
    \begin{subfigure}{0.32\textwidth}
        \includegraphics[width=\textwidth]{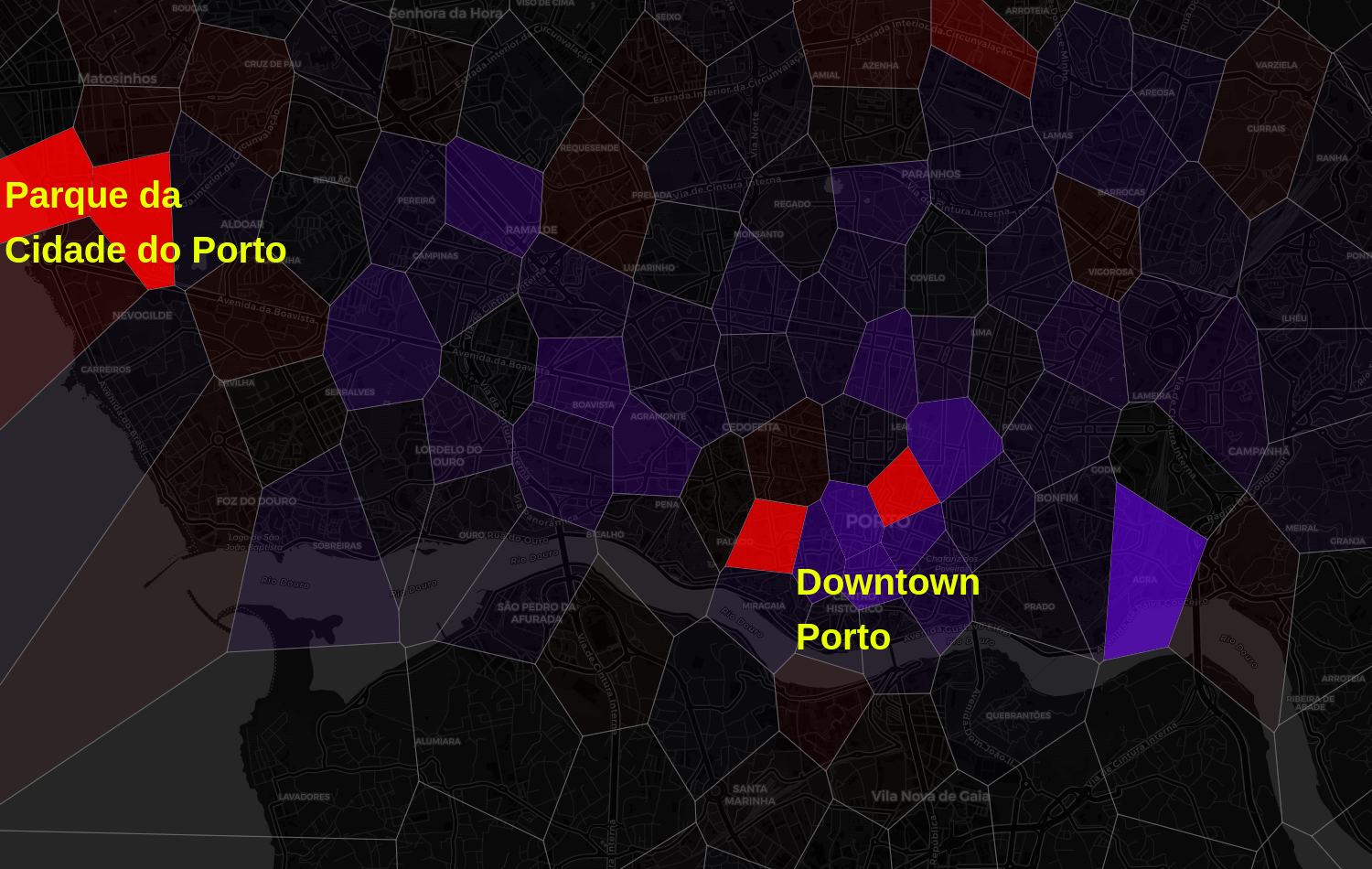}
        \caption{\label{fig:known_events:b}}
    \end{subfigure}
    \hfil
    \begin{subfigure}{0.32\textwidth}
        \includegraphics[width=\textwidth]{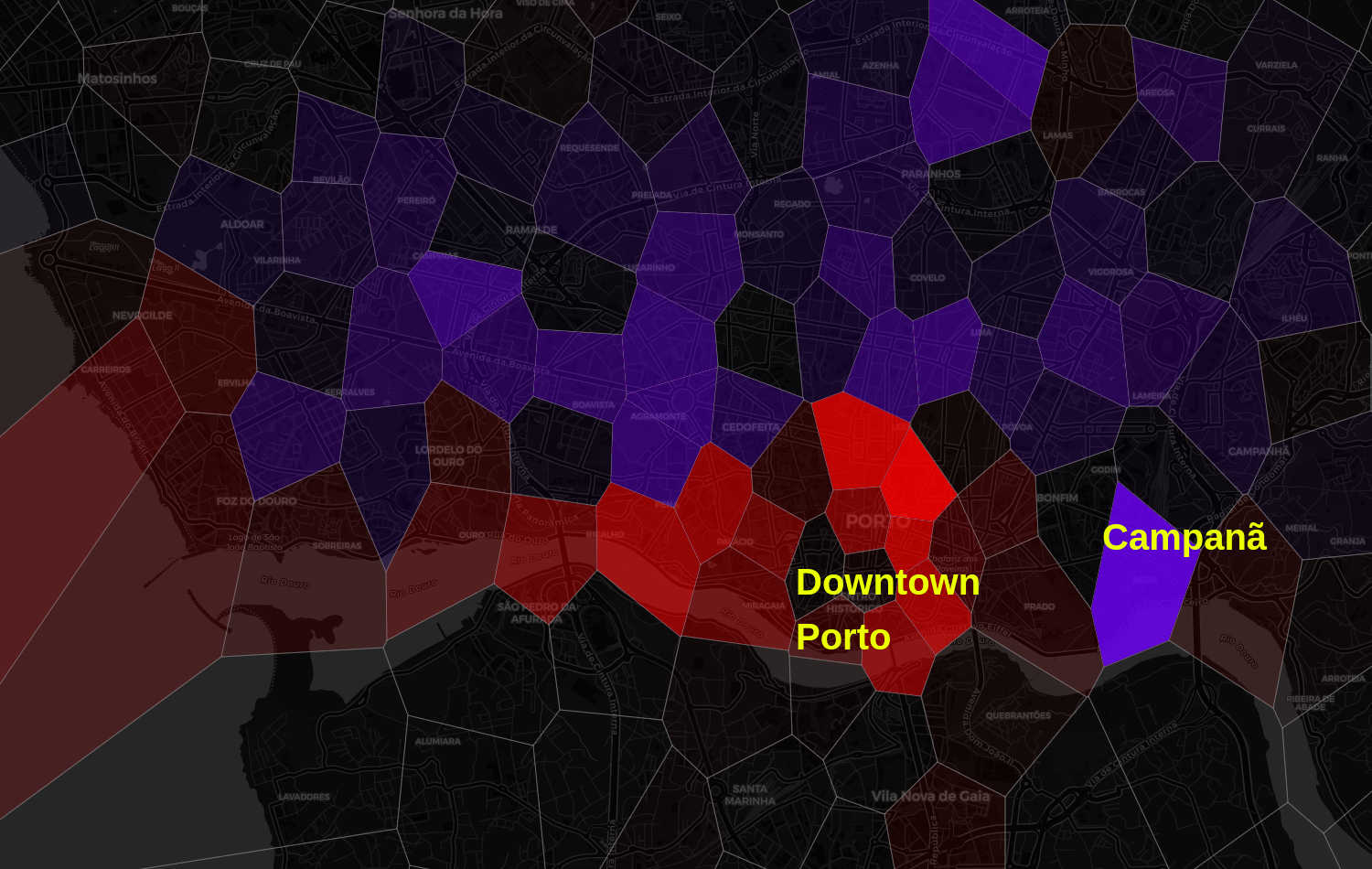}
        \caption{\label{fig:known_events:c}}
    \end{subfigure}
    \caption{Comparing the true taxi traffic $\mathbf{x}'$ with the best-matching unit's estimate, $\mathbf{w}'_b$, for a series of known events (or a single day selected from a multi-day event). \emph{Red}: more traffic than expected, \emph{blue}: less. \textbf{\protect\subref{fig:known_events:a}}: Behavioral shifts on New Year's Day 2014, part of a sustained period of slightly-depressed $a'$ during the winter holidays. Traffic increases at beaches in the west, decreases in Campanh{\~a} (a major local train station) in the east, and shifts within downtown. \textbf{\protect\subref{fig:known_events:b}}: Tuesday, May 6, 2014, a selection from \emph{Queima das Fitas} 2014 and the date of the \emph{Cortejo Acad{\'e}mico} (Academic Procession), the festival's peak. We see increases in traffic near \emph{Parque da Cidade do Porto} (Porto City Park) in the northwest; many of the celebrations take place adjacent to this park \cite{queimadasfitascitypark}. We also observe increases downtown near the parade's typical route \cite{queimadasfitasprocession}. \textbf{\protect\subref{fig:known_events:c}}: The night of June 23--24, 2014 is associated with a drop in activity in Campanh{\~a} and a large spike in taxi activity due to the \emph{Festa de S{\~a}o Jo{\~a}o do Porto} (St. John's festival), especially downtown and along the Douro river, where the festivities and fireworks take place, respectively \cite{festasaojoao}.}
    \label{fig:known_events}
\end{figure*}

Although our data come from GPS rather than visual data, the structure of GWR results in a close analogy to the early processing stages of biological vision (for a review, see \cite{spillmann2014receptive}). In biological vision, any given optic nerve fiber will only respond to inputs within a restricted sub-region of the overall visual field, known as that fiber's \emph{receptive field} \cite{hartline1938response}.  Evidence suggests that each receptive field maps to an equal amount of space in the brain's cortex, i.e., an equal amount of processing and subjective ``perception'' \cite{ransom1980perceptive, hubel1974uniformity}. However, each field represents a differently-sized \emph{input} region.  In the fovea (the visual field's center), many small receptive fields are packed closely together; in the periphery, larger fields are spread apart. The relationship between the size of a receptive field and its distance from the fovea is approximately linear \cite{hubel1974uniformity}. Because receptive fields are more densely packed in the fovea, and each field receives equal cortical processing, this structure gives rise to a subjective experience of detailed high-resolution visual experience in the center of the visual field, combined with blurry low-resolution peripheral vision.

In our system, the computation of the receptive fields is a Voronoi decomposition (\cref{fig:gwr_origins:c}).  Just as we can determine a best-matching unit $b$ given some input point $\mathbf{x}$, we can choose a region of points $\{\mathbf{x}\}$ for which a specific neuron $i$ would be the BMU,
\begin{equation}
   R_i := \{\mathbf{x} \mid  \lVert \mathbf{x} - \mathbf{w}_i \rVert ^ 2  \leq \lVert \mathbf{x} - \mathbf{w}_j \rVert ^ 2 \forall j \neq i \}
\end{equation}
which forms a partition of the input space.  In areas of the input space densely packed with neurons (e.g., downtown), receptive fields are quite small; conversely, in areas with a low density of neurons, receptive fields are larger---closely mimicking biological vision. The connection between GWR (and related methods) and Voronoi decompositions has long been key to their development \cite{fritzke1995growing, marsland2002self}. In practice, these Voronoi regions provide a better tokenization of the GPS input space than many popular alternatives: a set of regions which can be constructed incrementally, with guarantees that they (i) partition the full input space (any $\mathbf{x}$ can be assigned to a region), (ii) always are convex polyhedra, and (iii) adapt to the statistical density of the underlying data stream.  As in biological vision, we ensure that each receptive field receives equal processing in higher-level analysis regardless of its size, by treating each as a separate feature with equal scale (as described below).

\subsection{Modeling normal and anomalous dates}
\label{sec:modeling_dates}

\begin{figure*}
    \centering
    \begin{subfigure}{.32\textwidth}
        \includegraphics[width=\textwidth]{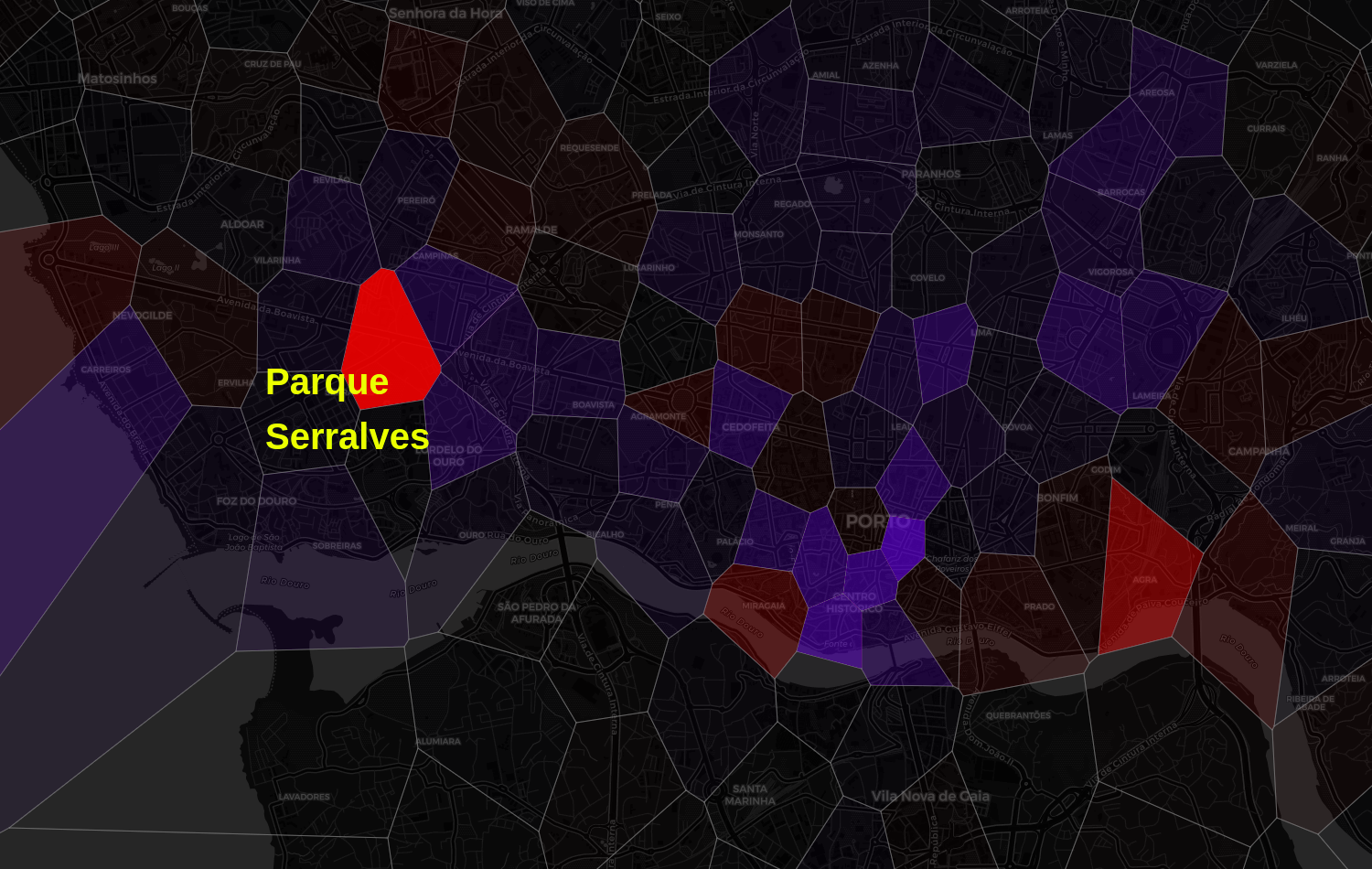}
        \caption{\label{fig:discovered_anomalies:a}}
    \end{subfigure}
    \hfil
    \begin{subfigure}{.32\textwidth}
        \includegraphics[width=\textwidth]{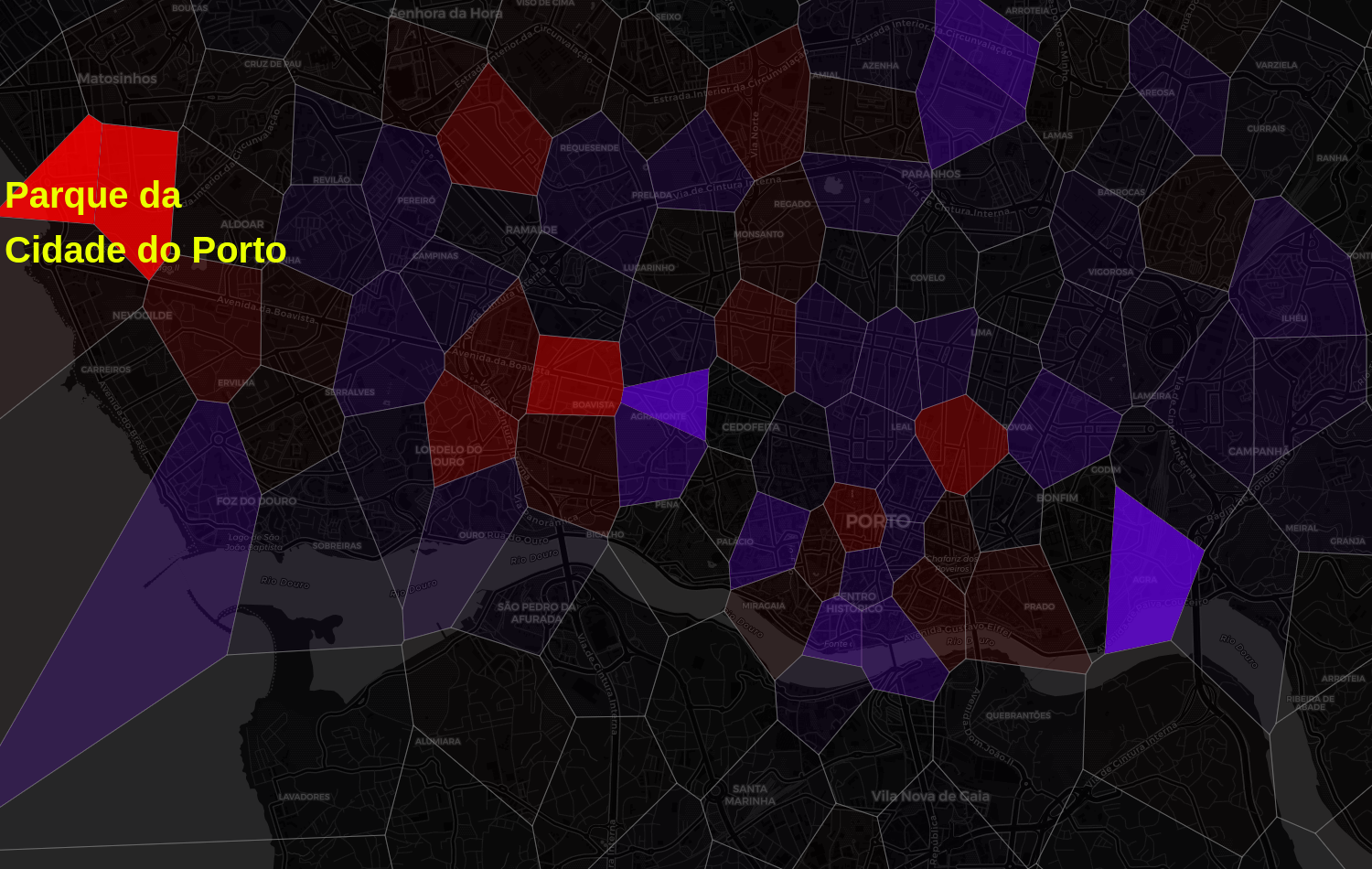}
        \caption{\label{fig:discovered_anomalies:b}}
    \end{subfigure}
    
    \caption{Comparing the true taxi traffic $\mathbf{x}'$ with the best-matching unit's estimate, $\mathbf{w}'_b$. \emph{Red}: more traffic than expected, \emph{blue}: less.  \textbf{\protect\subref{fig:discovered_anomalies:a}}: June 1, 2014. The citywide activity score $a'$ drops sharply to $a' = 0.883$, lower than any other date (other than the first training example provided to the model). A single region near \emph{Parque Serralves} (Serralves Park) has more than quadruple the taxis expected by the GWR, caused by a massive festival hosted there by a local arts and cultural institution. \textbf{\protect\subref{fig:discovered_anomalies:b}}: June 6, 2014. The citywide activity score $a'$ drops suddenly ($a' = 0.935$) due to a much greater-than-expected density of taxi traffic originating in two receptive fields near the Atlantic coast,  which comprise the \emph{Parque da Cidade do Porto} (Porto City Park). This anomaly is caused by the NOS Primavera Sound music festival.}
    \label{fig:discovered_anomalies}
\end{figure*}

As alluded to above, our anomaly-detection framework incorporates two levels of GWR.  In the second, we consider all trajectories $T_1, T_2, \ldots, T_D$ from a given day. Each one has a BMU $b$, a single neuron in the first-level GWR described above, implying an ordered list of neuron indices $b_1, b_2, \ldots, b_D$.

At the end of each day, we calculate a new input vector $\mathbf{x}'$ by finding the proportion of trajectories $T_i$ which occurred in each region. This forms the input of a second GWR. The Porto dataset, which is one year long, thus defines a sequence of 365 input vectors $\mathbf{x}'$, each of which represents a density of taxis originating in each region that day.  We represent the properties of this second GWR as $\mathbf{w}'_i$, $\eta'_i$, $\mathbf{E}'$, $a_T'$, etc. to distinguish them from the first GWR.  All hyperparameters can be found in the Appendix.

An unusual trait of this approach is that the size of the input $\mathbf{x}'$ can change, since it has one entry for each neuron in the first GWR, which is itself growing.  GWR implementations with input vectors of varying size are rare (although \cite{pitonakova2020robustness} is an exception). However, we note that if a new neuron (i.e., Voronoi region) is created in the first GWR at time $t$, then by assumption the previous traffic in that region should be 0 (it did not exist previously). Therefore, a simple implementation is to concatenate a 0 into previous memories for the new region, $\forall i~ \mathbf{w}'_i \leftarrow [\mathbf{w}'_i, 0]$.

In \cref{fig:citywide_patterns}, we visualize a selection of neurons from this higher-level GWR after training on the Porto dataset. Each neuron represents a prototype of an aggregated daily traffic distribution, and each actual observed day can be matched to a single prototype pattern, the best-matching unit $b'$. A large distance on day $t$ between an input $\mathbf{x}'$ and its BMU $\mathbf{w}'_b$ implies a low activity score $a'$. Thus, we can calculate this metric once per day and monitor it for any sudden drop in value, hinting at the existence of an unusual event or anomaly.

Based on our cultural research, we anticipated that we would find at least two major anomalous dates: (i) \emph{Queima das Fitas} (``Burning of the Ribbons''), an eight-day festival from May 4--11, 2014, and (ii) \emph{Festa de S{\~a}o Jo{\~a}o do Porto} (St. John's festival) on June 23, 2014. As can be seen in \cref{fig:activity_plot}, we can observe significant and sudden drops in the activity value $a'$ which coincide  with these events.  The activity score also is slightly depressed during the holiday season, from the weekend before Christmas through New Year's Day.  

The BMU weights $\mathbf{w}'_b$ represent a predicted relative taxi volume in each Porto region for the given day, which can be compared directly to the actual observed pattern $\mathbf{x}'$ to see which regions had more (or less) traffic than than expected. In \cref{fig:known_events}, we illustrate region-wise deviations from typical behavior in detail.

\subsection{Uncovering new events within Porto}
\label{sec:discovering_new_events}

 However, there are other drops in June 2014 which we did not anticipate, and which do not coincide with any official holiday. In \cref{fig:discovered_anomalies}, we show the results for this region-wise comparison on June 1 and June 6.  On June 1 (\cref{fig:discovered_anomalies:a}), we observe a massive surge in activity in a single Voronoi region near \emph{Parque Serralves} (Serralves Park), a 18-hectare park and part of a local art and cultural institution, \emph{Funda{\c{c}}{\~a}o de Serralves} (Serralves Foundation). This combined geospatial and temporal localization provides enough contextual clues to research what happened on that date: \emph{Serralves em Festa}, one of the largest festivals in Europe. The 2014 edition in particular marked the 25th anniversary of the Foundation and the 15th anniversary of its museum, \emph{Museu de Serralves}, leading to an especially significant celebration (archived at \cite{serralves} in Portugese).

In \cref{fig:discovered_anomalies:b}, we show a similar result for June 6, 2014. This anomaly can be traced to an unexpected surge of activity in two adjacent regions comprising \emph{Parque da Cidade do Porto} (Porto City Park) just south of the beaches of Matosinhos. In fact, researching this date and \emph{Parque da Cidade} uncovers contemporaneous reports of a major music festival that year, NOS Primavera Sound (archived at \cite{nosprimaverasound} in Portugese). (Note that the festival takes place June 5--7, but we model taxis \emph{originating} at regions, and empirically most traffic left the festival after midnight.  As a result, the anomalies are detected on June 6--8.)

\subsection{Evidence against catastrophic forgetting}

\begin{figure}
    \centering
    \includegraphics[width=.45\textwidth]{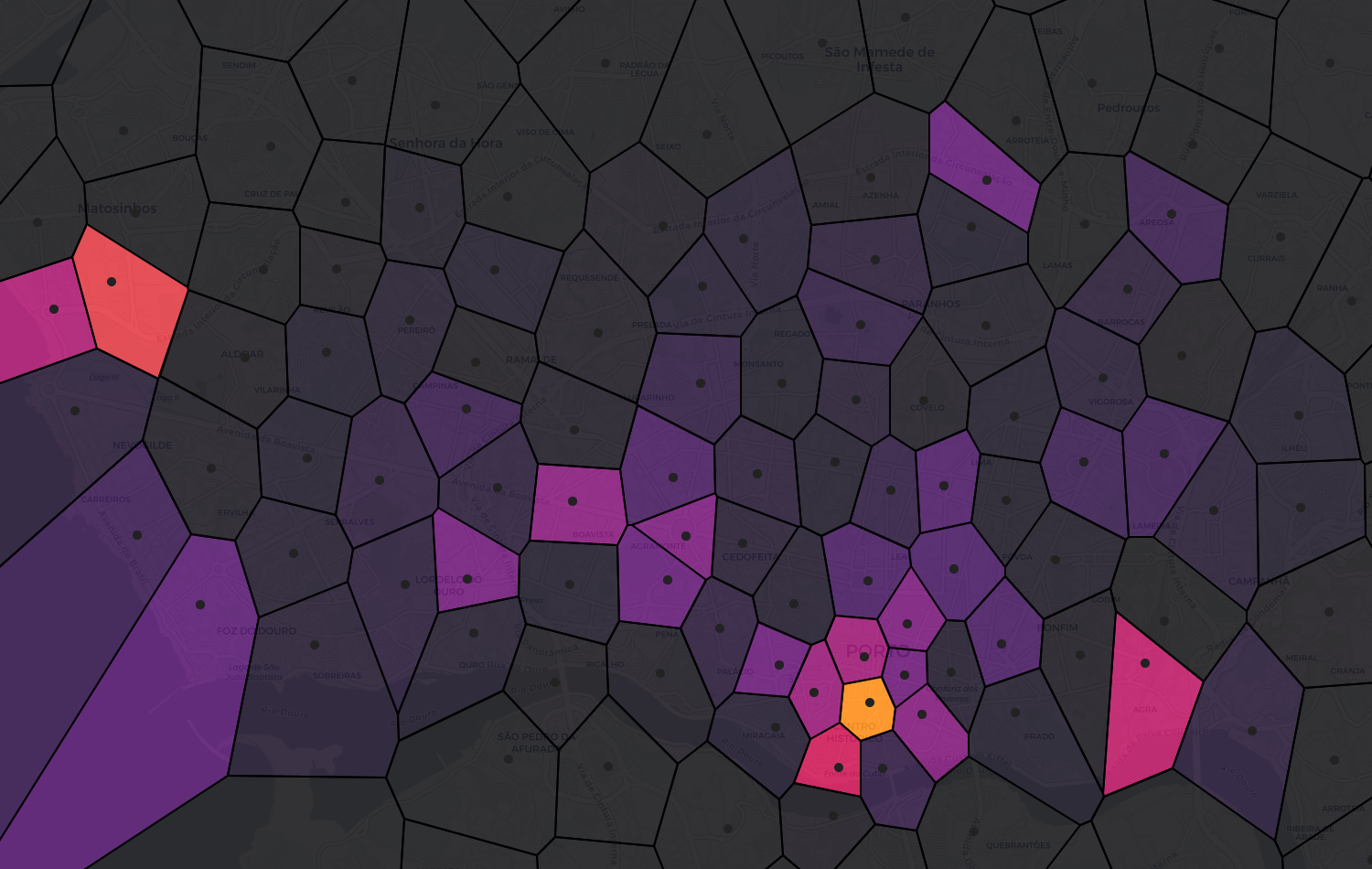}
    \caption{One of the final citywide neurons $\mathbf{w}'_i$ created by the second-level GWR, representing a much larger-than-otherwise-expected taxi density near \emph{Parque da Cidade do Porto} (Porto City Park) in the northwest. Because this neuron will be selected as the BMU $b$ for similar days, it absorbs many of the weight updates caused by the anomalous examples, protecting the rest of the network from catastrophic forgetting.}%
    \label{fig:citypark_neuron}
\end{figure}

Recall that lifelong learning problems such as this one, where knowledge must be acquired incrementally from a data stream, often struggle with catastrophic forgetting: the incorporation of new knowledge degrades information acquired in the past. In that case, the model would need to reacquire its old knowledge by redoing the training process, resulting in a slow increase in $a'$ mimicking the start of training (\cref{fig:activity_plot}, July 2013).  However, when anomalies are presented to a well-trained GWR, we observe that $a'$ instead returns to its high plateau quickly after the cessation of the unusual event (\cref{fig:activity_plot}, a--d). This remains true even when the unusual pattern is processed by the network many times in succession.  Consider the eight-day \emph{Queima das Fitas} festival (\cref{fig:activity_plot}a): the activity $a'$ drops suddenly at the beginning of the festival, but returns to a high value immediately afterward. If forgetting had occurred, we would observe a gradual return to higher performance as the model reacquired the lost knowledge.

This follows from two key properties of the GWR update rule (\cref{eq:eta_update,eq:w_update}).  First, updates are only applied to a BMU and its neighbors, $b \cup \mathcal{N}(b)$, preventing updates from propagating throughout the full network and restricting them to only the most relevant neurons. (The GWR synapse graph $\mathbf{E}$ tends to be relatively sparse, as visualized for the first-level GWR in \cref{fig:gwr_origins:b}). Second, the more often a neuron $i$ is activated, the lower its habituation $\eta_i$, and thus the smaller its weight update $\Delta \mathbf{w}_i$. Therefore, even when weight updates do propagate to a neuron, the network selectively preserves those which are already well-trained. 

In \cref{fig:citypark_neuron} we present additional evidence: one of the last neurons $\mathbf{w}'_i$ created by the second-level GWR. Repeated anomalies with high activity near  \emph{Parque da Cidade do Porto} (Porto City Park), such as \emph{Queima das Fitas} (\cref{fig:known_events:b}) and the NOS Primavera Sound music festival (\cref{fig:discovered_anomalies:b}) have created a newly-formed neuron, describing a traffic pattern with unusually-high taxi density in these two regions. Once this neuron is created, new inputs with high density in this area (e.g., subsequent days of \emph{Queima das Fitas}) match most closely with this neuron, making it the BMU $b$ and concentrating the weight updates on this neuron (\cref{eq:eta_update,eq:w_update}).  This protects other neurons from a poor weight update, automatically ``freezing'' their weights until normal traffic behavior resumes, as observed in \cref{fig:activity_plot}.

\section{Conclusion}

The challenges of catastrophic forgetting, concept drift, and model interpretability are difficult open problems in the machine learning community.  Where they intersect in the domains of geospatial and transportation modeling, we propose an unsupervised anomaly detector which incrementally learns patterns of life without catastrophic forgetting. These patterns are directly queryable by a human observer at any point in the training process, as are the anomalies themselves, and we show that the detector's results conform to both known and previously-unreported cultural events in the Porto area. We anticipate that applications of this anomaly detection framework will be of interest to numerous stakeholders with an interest in understanding human mobility patterns, including city governments, urban planners, transportation researchers, and other geospatial professionals.

\begin{acks}
This work was supported by NGA contract HM0476-21-C-0041. Approved for public release, 22-539.
\end{acks}

\bibliographystyle{ACM-Reference-Format}
\bibliography{paper}

\clearpage
\appendix

\section{Appendix}

\section*{GWR illustration on toy problem}

\begin{figure*}
    \centering
    \begin{subfigure}{0.32\textwidth}
        \includegraphics[width=\textwidth]{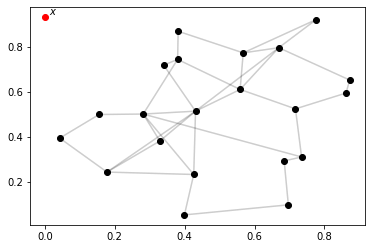}
        
        \caption{\label{fig:gwr_demo:a}}
    \end{subfigure}
    \hfil
    \begin{subfigure}{0.32\textwidth}
        \includegraphics[width=\textwidth]{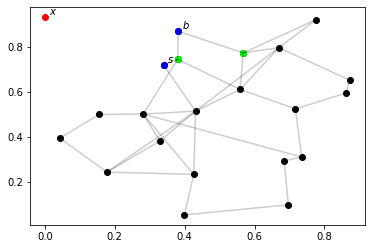}
        \caption{\label{fig:gwr_demo:b}}
    \end{subfigure}
    \hfil
    \begin{subfigure}{0.32\textwidth}
        \includegraphics[width=\textwidth]{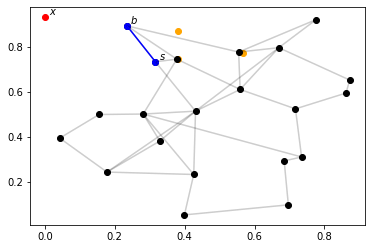}
        \caption{\label{fig:gwr_demo_c}}
    \end{subfigure}
    \caption{Illustrating the GWR algorithm on a toy problem, random points sampled from a uniform distribution $\mathcal{U}_{[0,1]}$. For this visualization, learning rates have been exaggerated for clarity. \textbf{\protect\subref{fig:gwr_demo:a}}: The state of the GWR network at the beginning of the training iteration. The neuron weight vectors $\{ \mathbf{w}_i \}$ are shown in black. A new point $\mathbf{x}$ (red) has been input to the network. \textbf{\protect\subref{fig:gwr_demo:b}}: the network finds the best-matching unit $b$ and second-best $s$ (blue). No edge exists yet between these neurons. Other neighbors of $b$, $\mathcal{N}(b)$, are shown in lime green. \textbf{\protect\subref{fig:gwr_demo_c}}: An edge is inserted between the neurons, and the neurons' weights are updated (with a larger learning rate for $b$ than $\mathcal{N}(b)$. For clarity, previous positions are shown in orange.}
    \label{fig:gwr_illustration1}
\end{figure*}

\begin{figure*}
    \centering
    \begin{subfigure}{0.32\textwidth}
        \includegraphics[width=\textwidth]{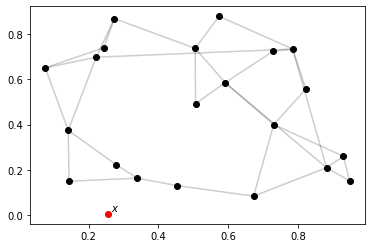}
        
        \caption{\label{fig:gwr_demo2:a}}
    \end{subfigure}
    \hfil
    \begin{subfigure}{0.32\textwidth}
        \includegraphics[width=\textwidth]{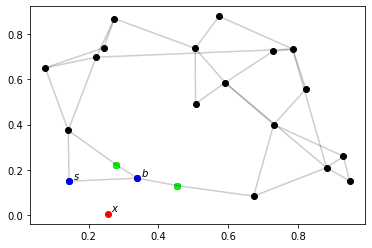}
        \caption{\label{fig:gwr_demo2:b}}
    \end{subfigure}
    \hfil
    \begin{subfigure}{0.32\textwidth}
        \includegraphics[width=\textwidth]{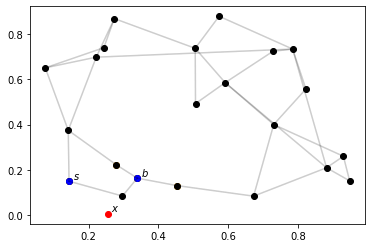}
        \caption{\label{fig:gwr_demo2_c}}
    \end{subfigure}
    \caption{Illustrating the GWR algorithm on a toy problem, random points sampled from a uniform distribution $\mathcal{U}_{[0,1]}$. For this visualization, learning rates have been exaggerated for clarity. Colors are the same as in \cref{fig:gwr_illustration1}. \textbf{\protect\subref{fig:gwr_demo2:a}}: The state of the GWR network at the beginning of the training iteration. \textbf{\protect\subref{fig:gwr_demo2:b}}: The network finds $b$ and $s$. An edge already exists here, which will be removed and replaced with a new neuron. \textbf{\protect\subref{fig:gwr_demo2_c}}: A new neuron is inserted between $\mathbf{x}$ and $b$, with connections to both $b$ and $s$. Additional edges may be adding during subsequent training iterations.}
    \label{fig:gwr_illustration2}
\end{figure*}

To provide a geometric intuition and visual illustration  of the GWR algorithm, as outlined in the main text, we include \cref{fig:gwr_illustration1,fig:gwr_illustration2} here.  We illustrate the algorithm on a toy problem, using random samples from a two-dimensional uniform distribution $\mathcal{U}_{[0,1]}$ with increased learning rates to better highlight the model's behavior in each iteration.

Figure \cref{fig:gwr_illustration1} demonstrates, at a high level, the GWR's behavior during a weight update (\cref{gwr_update}). Figure \cref{fig:gwr_illustration2} demonstrates, at a high level, the GWR's behavior during neurogenesis (\cref{gwr_neurogenesis}).

\section*{First-level GWR (regions)}

Note that these hyperparameters are largely the same as in previous work, although with different notational convention \cite{parisi2017lifelong, parisi2018lifelong}.  We make only narrow changes.  First, we alter the activity threshold $a_T$ to only add neurons when no existing neuron is closer than 1 km.  Second, we reduce the habituation rates $\tau_b, \tau_n$ to slow neurogenesis due to the immense number of trajectories (over 1 million) to be processed. Third, and finally, we disable neuron death (the removal of a neuron $i$ if $\mathcal{N}(i) = \varnothing$), because removing a neuron from the first-level GWR removes a learned region, deleting information from the second-level GWR that is essential for city-wide anomaly detection (\cref{fig:citywide_patterns}). At the end of training, this GWR contains 531 neurons total, each representing a region of Porto or the surrounding area.

\subsection*{Hyperparameter values}
\begin{tabular}{lll}
    $a_T$ & threshold for neurogenesis & $=\exp(-1)$ \\
    $f_T$ & threshold for neurogenesis & $=0.1$\\
    $\epsilon_b$ & learning rate (BMU) & $=0.5$\\
    $\epsilon_n$ & learning rate (neighbors) & $=0.005$ \\
    $\kappa$ & controls habituation & $=1.05$\\
    $\tau_b$ & controls habituation (BMU) & $=0.0133$ \\
    $\tau_n$ & controls habituation (neighbors) & $=0.00133$ \\
    $\mu_\textrm{max}$ & threshold for synapse pruning & $=200$ \\
\end{tabular}

\section*{Second-level GWR (city-wide patterns)}
The second-level GWR uses values previously published in the literature \cite{parisi2017lifelong, parisi2018lifelong} for all hyperparameters except $a_T'$, which we set to 1. As a result, neurogenesis is limited only by habituation $\eta_i'$, not by the activity $a'$, encouraging the formation of more neurons.  At the end of training, this GWR contains 69 neurons total, each representing a prototypical traffic pattern throughout Porto's regions.

\subsection*{Hyperparameter values}
\begin{tabular}{lll}
    $a_T'$ & threshold for neurogenesis & $=1$ \\
    $f_T'$ & threshold for neurogenesis & $=0.1$\\
    $\epsilon_b'$ & learning rate (BMU) & $=0.5$\\
    $\epsilon_n'$ & learning rate (neighbors) & $=0.005$ \\
    $\kappa'$ & controls habituation & $=1.05$\\
    $\tau_b'$ & controls habituation (BMU) & $=0.3$ \\
    $\tau_n'$ & controls habituation (neighbors) & $=0.1$ \\
    $\mu_\textrm{max}'$ & threshold for synapse pruning & $=200$ \\
\end{tabular}

\end{document}